\documentclass[journal]{IEEEtran}
\usepackage{graphicx}
\usepackage{multirow}
\usepackage{epstopdf}
\usepackage{times}
\usepackage{amsmath}
\usepackage{amssymb}
\usepackage{amsfonts}
\usepackage{color}
\usepackage{subfigure}
\usepackage{hyperref}

\usepackage{algorithm}
\usepackage{algorithmic}
\usepackage{rotating}
\usepackage{adjustbox,lipsum}
\usepackage{dsfont}
\usepackage{booktabs}
\usepackage{multirow}
\usepackage{array}

\usepackage{pifont}

\newcommand{\etal}{{\em et al.\,}}       
\newcommand{\eg}{{\em e.g.}}           
\newcommand{\ie}{{\em i.e.}}           

\begin{document}

\title{Label Distribution Learning for Generalizable Multi-source Person Re-identification}

\author{Lei Qi,
        Jiaying Shen,
        Jiaqi Liu,
        Yinghuan Shi,
        Xin Geng$^*$
\thanks{This work was supported in part by the National Key Research and Development Plan of China under Grant (2018AAA0100104), in part by Jiangsu Natural Science Foundation Project (BK20210224), the National Science Foundation of China (62125602, 62076063, 62222604 and 62192783), CAAI-Huawei MindSpore Project (CAAIXSJLJJ-2021-042A), China Postdoctoral Science Foundation Project (2021M690609) and CCF-Lenovo Bule Ocean Research Fund.}
\thanks{Lei Qi, Jiaying Shen, Jiaqi Liu and Xin Geng are with the School of Computer Science and Engineering, and the Key Lab of Computer Network and Information Integration (Ministry of Education), Southeast University, Nanjing, China, 211189 (e-mail: qilei@seu.edu.cn; shenjy@seu.edu.cn; liu\_jq@seu.edu.cn; xgeng@seu.edu.cn).}
\thanks{Yinghuan Shi is with the State Key Laboratory for Novel Software Technology, Nanjing University, Nanjing, China, 210023 (e-mail: syh@nju.edu.cn).}
\thanks{* Corresponding author: Xin Geng.}
}

%
%

\markboth{ }%
{Shell \MakeLowercase{\textit{et al.}}: Bare Demo of IEEEtran.cls for IEEE Journals}

\maketitle

\begin{abstract}
Person re-identification (Re-ID) is a critical technique in the video surveillance system, which has achieved significant success in the supervised setting. However, it is difficult to directly apply the supervised model to arbitrary unseen domains due to the domain gap between the available source domains and unseen target domains. In this paper, we propose a novel label distribution learning (LDL) method to address  generalizable multi-source person Re-ID task (\ie, there are multiple available source domains, and the testing domain is unseen during training), which aims to explore the relation of different classes and mitigate the domain-shift across different domains so as to improve the discrimination of the model and learn the domain-invariant feature, simultaneously. Specifically, during the training process, we produce the label distribution via the online manner to mine the relation information of different classes, thus it is beneficial for extracting the discriminative feature. Besides, for the label distribution of each class, we further revise it to give more and equal attention to the other domains that the class does not belong to, which can effectively reduce the domain gap across different domains and obtain the domain-invariant feature. Furthermore, we also give the theoretical analysis to demonstrate that the proposed method can effectively deal with the domain-shift issue. Extensive experiments on multiple benchmark datasets validate the effectiveness of the proposed method and show that the proposed method can outperform the state-of-the-art methods. Besides, further analysis also reveals the superiority of the proposed method.
\end{abstract}


%
\IEEEpeerreviewmaketitle

\section{Introduction}
\IEEEPARstart{P}{e}rson re-identification (Re-ID) is a significant technique for the public security, which can effectively improve the efficiency of capturing a specific person from the large-scale videos~\cite{DBLP:journals/tifs/ZhangXXLCFZ22}. Recently, person Re-ID has attracted an increasing interest in both academia and industry due to its great potential in the video surveillance application~\cite{zheng2016person,ye2020deep,DBLP:journals/tcsv/LengYT20,DBLP:journals/tifs/ZhuJYZSZ18}, which resorts to matching images of the same person captured by different cameras with the non-overlapping camera views. The main challenge of person Re-ID is the variations including body pose, viewing angle, illumination, image resolution, occlusion, background and so on across different cameras~\cite{DBLP:journals/pami/LiZG20,chen2021occlude,DBLP:journals/tcsv/QiWHSG20,DBLP:journals/tifs/MaJZTP20}. Generally, person Re-ID can be treated as a special case of the image retrieval problem with the goal of querying from a large-scale gallery set to quickly and accurately find images that match with a query image.


Currently, the typical person Re-ID methods have obtained excellent performance in the supervised setting due to the power of deep neural network~\cite{DBLP:journals/tmm/ZhaoLZZWM20,DBLP:journals/tmm/WeiZY0019,DBLP:conf/cvpr/WuZGL19,DBLP:conf/cvpr/LiZG18,DBLP:conf/cvpr/ZhengYY00K19,DBLP:journals/tomccap/QiWHSG21}. However, when these models are utilized to the unseen domains, the performance will drastically drop because of the data-distribution discrepancy between the available source domains and the unseen target domains. In general, when employing these methods in a new domain or scenario, we need to collect the data and give them labels, which is expensive and time-consuming, thus most of these existing supervised methods cannot be utilized in the real-world application. Although some unsupervised domain adaptation (UDA) methods are developed to mitigate the labeling task~\cite{DBLP:journals/tmm/YangYLJXYGHG21,DBLP:conf/cvpr/ZhaiLYSCJ020,DBLP:conf/iccv/WuZL19,DBLP:conf/eccv/ChenLL020,DBLP:conf/iccv/QiWHZSG19,DBLP:journals/tifs/LiCTYQ21,DBLP:journals/tifs/KhatunDSF21}, they still require to collect data and re-train the model for the new scenario. 

Domain generalization (DG) methods can address the above problem, which resorts to learning a model in the source domains and testing the model in the unseen domain~\cite{zhou2021domain}. In the person Re-ID community, some DG methods have been developed to obtain a robust model in the unseen target domain. For example, QAConv~\cite{DBLP:conf/eccv/LiaoS20} treats image matching as finding local correspondences in feature maps and constructs query-adaptive convolution kernels on the fly to achieve local matching. In~\cite{DBLP:conf/cvpr/ZhaoZYLLLS21}, a meta-learning strategy is introduced to simulate the train-test process of domain generalization for learning more generalizable models. RaMoE \cite{DBLP:conf/cvpr/DaiLLTD21} adopts an effective voting-based mixture mechanism to dynamically leverage the diverse characteristics of source domains to improve the generalization ability of the model. Differently, we aim at solving the issue from the perspective of label distribution learning (LDL)~\cite{DBLP:journals/tkde/Geng16}, 
which resorts to generating a label distribution for each class to promote the robustness of the model in the unseen domain.

In this paper, we focus on the generalizable multi-source person re-identification task as in~\cite{DBLP:conf/cvpr/ZhaoZYLLLS21}, where there are multiple available source domains in the training stage, and the testing data is unseen during training. To address this issue, we develop a novel label distribution learning method to enhance the discrimination and generalization of the model. To be specific, the label distribution of each class is generated using the online manner during the training course, which can accurately metric the similarity between different classes. Thus, this method can effectively mine the relation information of different classes to boost the discrimination of the model. Besides, to alleviate the discrepancy of data distribution across different domains, we further revise the label distribution of each class to give more and equal attention to the other domains that the class does not belong to. Thus, this scheme can help to learn the domain-invariant feature, which is beneficial for the model's generalization in the unseen domain. Moreover, we analyze the effectiveness of the proposed method from the theoretical perspective, which verifies that our method can indeed mitigate the domain gap across different domains. We conduct the experiments on multiple benchmark person Re-ID datasets to confirm the effectiveness of the proposed method. Moreover, the deep analysis by extensive experiments reveals the superiority of the proposed method.
In this paper, our main contributions can be summarized as:
  \begin{itemize}
    \item We develop a novel label distribution learning method for DG-Re-ID, which can not only explore the relation of different classes to boost the discrimination of the model but also reduce the domain gap across different domains to enhance the generalization ability of the model. 
    \item We propose a theoretical analysis to demonstrate the effectiveness of the proposed method, which shows that using the proposed label distribution learning can indeed map all samples into the same feature space and generate the domain-invariant feature.
    \item We evaluate our approach on multiple standard benchmark datasets, and the results show that our approach outperforms the state-of-the-art accuracy. Moreover, the ablation study and further analysis are provided to validate the efficacy of our method.
  \end{itemize}

The rest of this paper is organized as follows.
We review some related work in Section \ref{s-related}.
The proposed method is introduced in Section \ref{s-framework}.
The theoretical analysis is described in Section \ref{T_A}.
Experimental results and analysis are presented in Section \ref{s-experiment},
and Section \ref{s-conclusion} is conclusion.

\section{Related work}\label{s-related}
In this section, we review the most related works to our work, including the generalizable person re-identification, domain generalization and label distribution learning. The detailed investigation is presented in the following part.

\subsection{Generalizable Person Re-ID}
Person Re-ID methods have achieved a great success in computer vision in recent years. For example, in \cite{DBLP:journals/tip/ZhangGDWZC21}, a novel Deep High-Resolution Pseudo-Siamese Framework (PS-HRNet) is introduced to solve the matching problem of person with the same identity but different resolutions captured by different cameras, which can alleviate the difference of feature distributions between low-resolution images and high-resolution images. In order to address the cross-illumination person Re-ID task, Zhang~\etal~\cite{9761930} develop a novel Illumination Estimation and Restoring framework (IER), which can effectively reduce the disparities between training and testing images. Differently, the goal of domain generalizable person re-identification is to learn a robust model in the source domain that can directly perform well in the target domain without additional training. The existing methods mainly include network normalization, meta-learning and domain alignment. 

The network normalization methods study how to effectively combine Batch Normalization (BN) and Instance Normalization (IN)~\cite{DBLP:conf/nips/EomH19,DBLP:conf/bmvc/JiaRH19,DBLP:conf/cvpr/JinLZ0Z20}. For example, Jia~\etal~\cite{DBLP:conf/bmvc/JiaRH19} adopt this approach in Re-ID to eliminate the shifts in style and content of different domains by adding IN in specific layers. Since this method removes some discriminative information, Jin~\etal~\cite{DBLP:conf/cvpr/JinLZ0Z20} design a style normalization and restitution module to distill identity-relevant feature from the removed information and restitute it to the network to ensure high discrimination. Moreover, Choi~\etal~\cite{DBLP:conf/cvpr/ChoiKJPK21} alleviate the overfitting problem by investigating unsuccessful generalization scenarios with the help of batch-instance normalization.
Besides, some methods are designed based on meta-learning~\cite{DBLP:conf/icpr/LinCW20,DBLP:conf/cvpr/SongYSXH19,DBLP:conf/cvpr/ZhaoZYLLLS21,DBLP:conf/cvpr/DaiLLTD21}. For example, Song~\etal~\cite{DBLP:conf/cvpr/SongYSXH19} design a domain-invariant mapping network to generate classifier weights of specific categories to ensure good generalization performance on new datasets with the help of meta-learning. In~\cite{DBLP:conf/cvpr/ZhaoZYLLLS21}, the memory-based multi-source meta-learning framework is introduced to enable the model to simulate the train-test process of DG during training and diversify meta-test feature with a meta batch normalization layer. 


The main idea of domain alignment is to map all samples from different domains into the same space to alleviate the discrepancy of data distribution~\cite{DBLP:conf/aaai/ChenDLZX0J21,DBLP:conf/eccv/LuoSZ20,DBLP:conf/wacv/YuanCCYRW020,DBLP:conf/eccv/ZhuangWXZZWAT20,DBLP:conf/eccv/LiaoS20}. In~\cite{DBLP:conf/aaai/ChenDLZX0J21}, a dual distribution alignment network is proposed to map images into a domain-invariant feature space by selectively aligning the distributions of multiple source domains. 
Yuan \etal~\cite{DBLP:conf/wacv/YuanCCYRW020} employs an adversarial domain-invariant feature learning framework to learn separate identity-related feature from challenging variations, using video timestamp and camera index. 
 Furthermore, QAConv~\cite{DBLP:conf/eccv/LiaoS20} treats image matching as finding local correspondences in feature maps and constructs query-adaptive convolution kernels on the fly to achieve local matching. 
In addition to all the aforementioned methods, Liao~\etal~\cite{liao2021graph} propose to explore the use of hard example mining in the data sampling stage, which builds a nearest neighbor relationship graph for all classes to provide informative and challenging examples for learning.

In this paper, we deal with the generalizable person Re-ID task from the perspective of label distribution learning. On the one hand, our proposed method aims to explore the relation of different classes. On the other hand, our method can also alleviate the data-distribution discrepancy across different domains to learn domain-invariant feature simultaneously.

\subsection{Domain Generalization}

Recently, some methods are also developed to address the domain generalization problem in the classification and semantic segmentation tasks~\cite{DBLP:conf/cvpr/NamLPYY21,DBLP:conf/eccv/SeoSKKHH20,DBLP:conf/iccv/YueZZSKG19,DBLP:conf/cvpr/CarlucciDBCT19,DBLP:conf/nips/BalajiSC18,DBLP:conf/iccv/LiZYLSH19,DBLP:conf/eccv/LiTGLLZT18,DBLP:journals/pr/ZhangQSG22}.
Inspired by domain adaptation methods, some works based on domain alignment~\cite{DBLP:conf/eccv/LiTGLLZT18,DBLP:conf/nips/ZhaoGLFT20,DBLP:conf/icml/MuandetBS13,DBLP:journals/pr/RahmanFBS20,DBLP:conf/cvpr/LiPWK18,DBLP:conf/cvpr/GongLCG19,DBLP:conf/wacv/RahmanFBS19} resort to mapping all data from different domains into the same space to alleviate the difference of data distribution across different domains. For example, Muandet~\etal~\cite{DBLP:conf/icml/MuandetBS13} propose a kernel based optimization algorithm to learn the domain-invariant feature and enhance the generalization ability of the feature representation. However, this method cannot ensure the consistency of conditional distribution, hence Zhao~\etal~\cite{DBLP:conf/nips/ZhaoGLFT20} introduce an entropy regularization term to measure the dependency between the learned feature and the class labels, which can effectively ensure the conditional invariance of learned feature, so that the classifier can also correctly classify the feature from different domains. 

 Besides, Gong~\etal~\cite{DBLP:conf/cvpr/GongLCG19} utilize CycleGAN~\cite{DBLP:conf/iccv/ZhuPIE17} to generate new styles of images that cannot be seen in the training data, which smoothly bridge the gap between source and target domains to improve the generalization of the model. Rahman~\etal~\cite{DBLP:conf/wacv/RahmanFBS19} also use GAN to generate synthetic data and then reduce domain discrepancy to achieve domain generalization. Li~\etal~\cite{DBLP:conf/cvpr/LiPWK18} adopt an adversarial autoencoder learning framework to learn a generalized latent feature representation in the hidden layer, and use Maximum Mean Discrepancy to align source domains, then they match the aligned distribution to an arbitrary prior distribution via adversarial feature learning. In this way, it can better generalize the feature of the hidden layer to other unknown domains. Rahman~\etal~\cite{DBLP:journals/pr/RahmanFBS20} incorporate the correlation alignment module along with adversarial learning to help achieving a more domain agnostic model due to the improved ability to more effectively reduce domain discrepancy.
 In addition to performing adversarial learning at the domain level to achieve domain alignment, Li~\etal~\cite{DBLP:conf/eccv/LiTGLLZT18} also perform domain adversarial tasks at the 
class level to align samples of each category that from different domains.

Particularly, our proposed method is related to the domain alignment methods. Differently, we achieve this goal by generating a special label distribution, which is not well investigated in the existing works. Moreover, since the person Re-ID is a metric task (\ie, this goal is to identify the same or different persons, which is different from the typical classification task), therefore it is also excellently significant to mine the relation information of different classes in this task.

\subsection{Label Distribution Learning}
Label distribution learning (LDL) is proposed for some applications with label ambiguity~\cite{DBLP:journals/tkde/Geng16}, such as emotion distribution learning, age estimation, sense beauty, classification tasks, etc. LDL allows direct modeling of different importance of each label to the instance, and thus can better match the nature of many real applications. For example, Chen~\etal~\cite{DBLP:conf/cvpr/ChenWCSGR20} propose a label distribution learning on auxiliary label space graphs to  address the problem that some facial expression recognition datasets only contain one-hot labels instead of label distributions. Besides, Huo~\etal~\cite{DBLP:conf/cvpr/HuoYXZHLG16} put forward a deep age distribution learning which generates a Gaussian age distribution for each facial image as the training target and uses ensemble method to get the result.

Different from these label distribution learning methods, we resorts to developing the label distribution learning to mine the relation information of different classes so as to improve the discrimination of the model. In addition, we also expect that the proposed LDL can mitigate the data-distribution discrepancy across different domains to learn the domain-invariant feature.

\section{The proposed method}\label{s-framework}

\begin{figure*}[t]
\centering
\includegraphics[width=16cm]{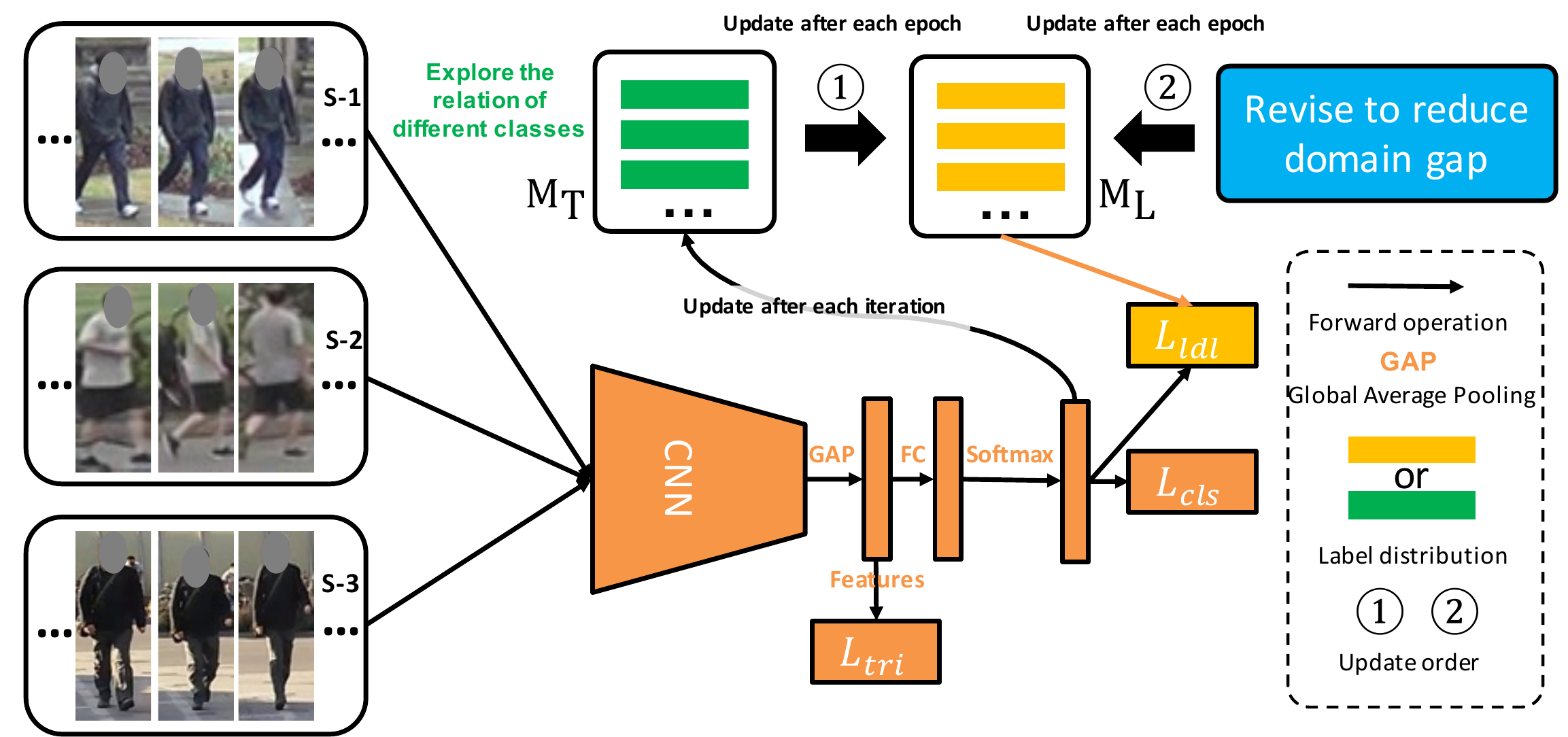}
\caption{An illustration of the proposed LDL method. Here we take three source domains as an example. As seen in this figure, our method yields the label distribution set $\mathbf{M_{L}} \in \mathbb{R}^{C \times C}$ via the step-1 and step-2 after each epoch, which is leveraged to train the model. It is worth noting that the $i$-th row in $\mathbf{M_{L}}$ is the label distribution of the $i$-th class. Best viewed in color.}
\label{fig03}
\end{figure*}

In this paper, to enhance the generalization capacity of the model to the unseen domain, we propose to employ the label distribution learning for generalizable multi-source person re-identification (Re-ID). Particularly, the proposed label distribution is designed for mining the relation information of different classes and reducing the domain gap across different domains, simultaneously. The pipeline of our method is illustrated in Fig.~\ref{fig03}. In the following part, we will detailedly introduce the process of generating the label distribution.

\subsection{Exploring the Relation of Different Classes} \label{LDL-1}
Most of the generalizable person Re-ID methods neglect the relation of different classes, which can help the model to improve the feature discrimination. Specifically, because the person Re-ID is the metric task (\ie, the goal of this task is to match the feature to identify the same or different persons), the feature discrimination is excellently significant in the person Re-ID task. In the generalizable Re-ID task, the training set consists of multiple source domains, which contain a quantity of classes, thus exploring the relation of different classes is beneficial to improve the performance of the model. In this paper, we achieve this goal using label distribution learning, 
therefore the design of the label distribution is very critical in our method.

To obtain the similarity across different classes, we directly utilize the output of the classier (\ie, softmax layer) in the neural network. For example, for a sample from the $i$-th class, if it is more similar to the $j$-th class than the other classes, the classifier's prediction should be larger in the $j$-th class than in the other classes except for itself, which has been validated in the literature \cite{DBLP:conf/cvpr/WuXYL18}. Based on the fact, we build the relation of different classes using the classifier's output. Specifically, we first initialize two similarity matrix $\mathbf{M_T}$ and $\mathbf{M_L} \in \mathbb{R}^{C \times C}$, where $C$ is the total number of the classes and each element in the similarity matrix means the similarity of two corresponding classes. For each training image, we can obtain its prediction (\ie, the output of the softmax layer), and then update the corresponding row in $\mathbf{M_T}$ by
\begin{equation}
  \begin{aligned}
  \mathbf{M_T}[i, :] \leftarrow  (1-m)\mathbf{M_T}[i, :] + m P_i, 
  \end{aligned}
  \label{eq01}
\end{equation}
where $ P_i$ is the prediction of classifier for the samples from the $i$-th class, and $m$ is the hyper-parameter of a momentum update. After finishing an epoch, we update $\mathbf{M_L}$ by $\mathbf{M_T}$. Finally, $\mathbf{M_L}$ is employed as label distribution set to explore the relation of different classes, which can bring more discriminative features for the unseen domain. It is worth noting that the $i$-th row in $\mathbf{M_L}$ indicates the label distribution for the $i$-th class. Besides, we initialize $\mathbf{M_L}$ and $\mathbf{M_T}$ as $\frac{1}{C}$ in our method.



\subsection{Reducing the Domain Gap across Different Domains}\label{LDL-2}
In this part, we will introduce how to reduce the domain gap across different domains via further processing the above label distribution set (\ie, $\mathbf{M_L}$). To better alleviate the domain gap, during the process of the label distribution generation, we require to give more attention to the cross-domains than the domain that a class belongs to. Besides, to better align the data distribution across different domains, we aim to pull all classes closer to the far cross-domains, which can enhance the generalization ability of the model in the unseen domain via yielding the domain-invariant feature for all domains.

For the label distribution of all samples from the $i$-th class (\eg, $L_i =\mathbf{M_L}[i, :] \in \mathbb{R}^{1 \times C}$), we set these classes from the same domain as the $i$-th class to $0$ except for the $i$-th value. Particularly, the $i$-th value remains unchanged. Here, we define $\mathbf{I}_d$ as the $d$-th domain set of the class index in $L_i$, thus we can compute the similarity between the $i$-th class and the $d$-th domain as follows:
  \begin{equation}
  \begin{aligned}
  s_i^d = \frac{1}{N_d}\sum_{j=1}^{N_d} L_i[\mathbf{I}_d(j)],~~s.t., d \neq \phi(i),
  \end{aligned}
  \label{eq02}
  \end{equation}
  where $N_d$ is the number of classes (\ie, IDs) in the $d$-th domain. $\phi(\cdot)$ is the function that maps a sample to its domain, \ie, we can know which domain a sample belongs to via $\phi(\cdot)$. From the perspective of data distribution, the small value indicates that the distance between the $d$-th domain and all samples from the $i$-th class is far. To address this issue, we use the averaged value to re-assign the similarity of the $d$-th domain and the $i$-th class as bellow:

    \begin{equation}
  \begin{aligned}
  \bar{s_i} = \frac{1}{K-1}\sum_{d\in \{\mathbf{D} \setminus  \phi (i) \}} s_i^d,~~s.t., d \neq \phi(i),
  \end{aligned}
  \label{eq03}
  \end{equation}
  where $K$ is the number of domains in the training set, and $\mathbf{D}$ is the set of all domains. $\mathbf{D} \setminus  \phi (i)$ denotes the set excluding $\phi (i)$ from $\mathbf{D}$. It is worth noting that the averaged value for all domains makes the large (small) value become small (large) value, thus this can give more attention to the domain that have the long distance from a class, and few attention to the domain that have short distance from a class.

  Here, let the label distribution of the $i$-th class as $L_i=[\cdots, l_i, \cdots, l_j, \cdots]$. We assign the new value to $l_j$ as:

    \begin{equation}
  \begin{aligned}
     l_j=\frac{l_j}{s_i^{\phi(j)} \times N_{\phi(j)}} \times & \frac{(1-l_i) \times \bar{s_i} \times N_{\phi (j)}}{\sum_{d\in \{\mathbf{D} \setminus  \phi (i) \}} \bar{s_i} \times N_d} \\=\frac{l_j}{s_i^{\phi(j)} } \times & \frac{1-l_i}{\sum_{d\in \{\mathbf{D} \setminus  \phi (i) \}}  N_d}, \\
    & s.t., \phi (j) \neq \phi (i),
  \end{aligned}
  \label{eq04}
  \end{equation}
  thus we can guarantee the sum of the label distribution is $1$. 

\subsection{The Training Process}
In this paper, we use the following loss to conduct the label distribution learning as:
   \begin{equation}
  \begin{aligned}
  L_{ldl} = -\sum_{i=1}^{C} l_i\times log(p_i), 
  \end{aligned}
  \label{eq05}
  \end{equation}
where $C$ is the total number of classes in the training set, and $p_i$ represents the output probability of a sample from the $i$-th class.

During training, we employ the cross-entropy loss (\ie, $L_{cls}$), the triplet loss (\ie, $L_{tri}$) with hard mining sampling~\cite{ulyanov2016instance} and the label distribution learning (\ie, $L_{ldl}$)  to train the model. Particularly, the cross-entropy loss and the triplet loss are the basic loss in the person Re-ID community~\cite{DBLP:journals/tmm/LuoJGLLLG20,DBLP:conf/iccv/FuWWZSUH19}. The overall loss for training the model can be described as:
  \begin{equation}
  \begin{aligned}
  L_{overall}=L_{cls}+L_{tri}+\lambda L_{ldl},
  \end{aligned}
  \label{eq06}
  \end{equation}
  where $\lambda$ is the hyper-parameter to trade off the basic loss and the label distribution learning.
The overall process of the label distribution is described in Algorithm~\ref{al01}.

\textit{Remark:} The generated label distribution can also be considered as soft-label, \ie, there are several non-$0$ values in the label of each class. Besides, it is worth noting that the label smoothing~\cite{DBLP:conf/cvpr/SzegedyVISW16} scheme is also a kind of label distribution, which is usually utilized to promote the model's robustness. However, the scheme does not explore the relation of different classes and reduce the domain gap across different domains. Particularly, we also use the cross-entropy loss with the label smoothing scheme in all experiments.

\begin{algorithm}[t]
\caption{\small{The training process of the proposed LDL}}~\label{alg1}
\begin{algorithmic}[1]
\STATE {\bf Input}: Training examples $\mathrm{X}$ and labels $\mathrm{Y}$.
\STATE {\bf Output}: The final model's parameter  $\theta$.
\STATE {\bf procedure} 
LDL 
\STATE $\theta \leftarrow$ Initialize by ResNet-50 pre-trained on ImageNet.\\
\STATE $\mathbf{M_L}$ and $\mathbf{M_T}$ $\leftarrow$ Initilize by $\frac{1}{C}$.
\FOR{epoch $\in [1,...,T]$}
\FOR{iteration $\in [1,...,T^{'}]$}
\STATE $\mathbf{M_T}[i, :] \leftarrow  (1-m)\mathbf{M_T}[i, :] + m P_i$.
\STATE Update the parameter $\theta$ by Eq.~\ref{eq06}.
\ENDFOR
\STATE $\mathbf{M_L}$ $\leftarrow$ Update by $\mathbf{M_T}$.
\STATE For the $i$-th class LD, if a class is in the same domain as the $i$-th class, it is set as 0. 
\STATE $\mathbf{M_L}$ $\leftarrow$ Update by Eq.~\ref{eq04}.
\ENDFOR
\STATE {\bf end procedure}
\end{algorithmic}
\label{al01}
\end{algorithm}

\section{Theoretical Analysis}\label{T_A}
In this part, we will give the theoretical analysis to validate that the proposed label distribution learning can indeed learn the domain-invariant feature well (\ie, our method can effectively reduce the discrepancy of the data distribution across different domains).

\textbf{Proposition.} \textit{Let ${\mathcal S_i^k}$ and ${\mathcal S_j^d}$ denote samples from the $i$-th (or $j$-th) class of the $k$-th (or $d$-th) domain. $p_i^k(x)$ and $p_j^d(x)$ are their probability density functions. It can be proved that ideally, using the proposed label distribution learning will lead to}
\begin{eqnarray}\label{eqn:proposition} 
p_i^k(x) &=& p_j^d(x), \quad \forall x \in {\mathcal S_i^k}\cup{\mathcal S_j^d}.
\end{eqnarray}

\textbf{Proof.} All the following analysis is conducted in the context of the learned feature representation (or equally, the learned shared subspace). Given an image $x^{d}_{i}$ from the $i$-th class in the $d$-th domain, its posterior probability with respect to the $k$-th domain (denoted by $\mathcal{D}^k (k\in \{1,\cdots,D\})$) can be expressed via the Bayes' rule as 
\begin{equation}\label{eqn:Bayes}
P(\mathcal{D}^k|x^d_i) = \frac{p(x^d_i|\mathcal{D}^k)P(\mathcal{D}^k)}{p_i^d(x^d_i)},\quad \forall k \in \{1,\cdots,D\},
\end{equation}where $p(x^d_i|\mathcal{D}^k)$ is the class-conditional probability density function in the $k$-th domain, $p_i^d(x^d_i)$ denotes the probability density function of the images in the $i$-th class of the $d$-th domain, and $P(\mathcal{D}^k)$ is the prior probability of the $k$-th domain. 


Let us turn to Eq.~\ref{eqn:Bayes} and rearrange it as
\begin{equation}\label{eqn:Bayes-1}
p(x^d_i|\mathcal{D}^k) = \frac{P(\mathcal{D}^k|x^d_i)}{P(\mathcal{D}^k)}p^d_i(x^d_i),\quad \forall k \in \{1,\cdots,D\},
\end{equation}Without loss of generality, equal prior probability can be set for the $D$ domains, that is, $P(\mathcal{D}^k)$ is constant $1/D$. Further, note that by optimizing $D_j$ in Eq.~\ref{eq05} above, it can be known that 
\begin{equation}\label{eqn:Bayes-1-1}
P(\mathcal{D}^k|x^d_i)=
\left\{\begin{matrix}
 \frac{1-l^d_i}{D-1} & k \neq d\\ 
 l^d_i &  k = d\\ 
\end{matrix}\right.
\end{equation}
Combining the above results, Eq.~\ref{eqn:Bayes-1} becomes
\begin{equation}\label{eqn:Bayes-2}
\begin{aligned}
p(x^d_i|\mathcal{D}^k) =\left\{\begin{matrix}
 \frac{(1-l_i^d)\times D}{D-1}p_i^d(x^d_i) & k \neq d\\ 
 l_i^d \times p_i^d(x^d_i) \times D & k=d
\end{matrix}\right.
\end{aligned}
\end{equation}

Assuming there are $3$ (\ie, $D=3$) source domains in the training stage, we have 

\begin{equation}\label{eqn:Bayes-4}
\begin{aligned}
&p_j^1(x^2_i) = \sum_{k=1}^{3}p(x^2_i|\mathcal{D}^k)P(\mathcal{D}^k) \\
 &=p(x^2_i|\mathcal{D}^1)P(\mathcal{D}^1)+p(x^2_i|\mathcal{D}^2)P(\mathcal{D}^2)+p(x^2_i|\mathcal{D}^3)P(\mathcal{D}^3)\\
 &=D\times l_i^2 \times p_i^2(x^2_i)\times \frac{1}{D}+ \frac{D\times (1-l_i^2)}{D-1}\times p_i^2(x^2_i)\times \frac{2}{D}\\
 &=l_i^2 \times p_i^2(x^2_i)+\frac{2\times (1-l_i^2)}{D-1}\times p_i^2(x^2_i)\\
 &=l_i^2 \times p_i^2(x^2_i)+(1-l_i^2) \times p_i^2(x^2_i)=p_i^2(x^2_i).
\end{aligned}
\end{equation}
Similarly, the result for any given image $x^1_j$ from the $j$-th in the $1$-st domain can be obtained as
\begin{equation}\label{eqn:Bayes-4}
\begin{aligned}
&p_i^2(x^1_j) = \sum_{k=1}^{3}p(x^1_j|\mathcal{D}^k)P(\mathcal{D}^k) \\
 &=p(x^1_j|\mathcal{D}^1)P(\mathcal{D}^1)+p(x^1_j|\mathcal{D}^2)P(\mathcal{D}^2)+p(x^1_j|\mathcal{D}^3)P(\mathcal{D}^3)\\
 &=D\times l_j^1 \times p_j^1(x^1_j)\times \frac{1}{D}+ \frac{D\times (1-l_j^1)}{D-1}\times p_j^1(x^1_j)\times \frac{2}{D}\\
 &=l_j^1 \times p_j^1(x^1_j)+\frac{2\times (1-l_j^1)}{D-1}\times p_j^1(x^1_j)\\
 &=l_j^1 \times p_j^1(x^1_j)+(1-l_j^1) \times p_j^1(x^1_j)=p_j^1(x^1_j).
\end{aligned}
\end{equation}
Therefore, for the samples from any two domains, we can obtain as
\begin{equation}\label{eqn:Bayes-4}
\begin{aligned}
p_i^k(x) =p_j^d(x),  \quad \forall x \in {\mathcal S_i^k}\cup{\mathcal S_j^d},
\end{aligned}
\end{equation}
where $\mathcal S_j^d$ denotes all samples from the $j$-th class of the $d$-th domain. This means that the two distributions, $p_i^k(x)$ and $ p_j^d(x)$, are identical on the set ${\mathcal S_i^k}\cup{\mathcal S_j^d}$. With respect to the definitions of the two distributions, this indicates that \textit{upon the learned feature representation, the data distributions of any two domains become identical and the distribution discrepancy is therefore removed}. \quad\quad\quad\quad\quad\quad\quad\quad\quad\quad\quad\quad\quad\quad\quad$\blacksquare$

\section{Experiments}\label{s-experiment}
\renewcommand{\cmidrulesep}{0mm} 
\setlength{\aboverulesep}{0mm} 
\setlength{\belowrulesep}{0mm} 
\setlength{\abovetopsep}{0cm}  
\setlength{\belowbottomsep}{0cm}

In this part, we firstly introduce the experimental datasets and settings in Section~\ref{sec:EXP-DS}. Then, we compare the proposed method with the state-of-the-art generalizable Re-ID methods in Section~\ref{sec:EXP-CUA}, respectively. To validate the effectiveness of various components in the proposed framework, we conduct ablation studies in Section~\ref{sec:EXP-SS}. Lastly, we further analyze the property of the proposed method in Section~\ref{sec:EXP-FA}.
\subsection{Datasets and Experimental Settings}\label{sec:EXP-DS}
\subsubsection{Datasets} 
We evaluate our approach on four large-scale image datasets: Market1501~\cite{DBLP:conf/iccv/ZhengSTWWT15}, DukeMTMC-reID~\cite{DBLP:conf/iccv/ZhengZY17}, MSMT17~\cite{wei2018person} and CUHK03-NP~\cite{DBLP:conf/cvpr/LiZXW14,DBLP:conf/cvpr/ZhongZCL17}. 
 \textbf{Market1501 (Ma)} contains 1,501 persons with 32,668 images from six cameras. Among them, $12,936$ images of $751$ identities are used as a training set. For evaluation, there are $3,368$ and $19,732$ images in the query set and the gallery set, respectively. \textbf{DukeMTMC-reID (D)} has $1,404$ persons from eight cameras, with $16,522$ training images, $2,228$ query images and $17,661$ gallery images.
 \textbf{MSMT17 (Ms)} 
 is collected from a 15-camera network deployed on campus. The training set contains $32,621$ images of $1,041$ identities. For evaluation, $11,659$ and $82,161$ images are used as query and gallery images, respectively. \textbf{CUHK03-NP (C)} has an average of 4.8 images per camera for each identity. The dataset provides both manually labeled bounding boxes and DPM-detected bounding boxes. On this dataset, there are $7,365$ training images, and $1,400$ images and $5,332$ images in query set and gallery set are used in the testing stage. Particularly, we divide these four datasets into two parts: three domains as source domains for training and the other one as target domain for testing. We adopt the recommended setting in ~\cite{DBLP:conf/cvpr/ZhaoZYLLLS21}.
   For all datasets, we employ CMC (\ie, Cumulative Match Characteristic) accuracy and mAP (\ie, mean Average Precision) for Re-ID evaluation~\cite{DBLP:conf/iccv/ZhengSTWWT15}.

\subsubsection{Implementation Details} 
In this experiment, we use the ResNet-50~\cite{DBLP:conf/cvpr/HeZRS16} and IBN-Net50~\cite{DBLP:conf/eccv/PanLST18} pre-trained on ImageNet~\cite{DBLP:conf/cvpr/DengDSLL009} to initialize the network parameters. For the cross-entropy loss, we employ the label smoothing scheme during the training course. In a batch, the number of IDs and the number of images per person are set as $16$ and $4$ to produce triplets for each domain, respectively.
 The initial learning rate is $3.5\times 10^{-4}$ and divided by $10$ at the $30$-th and $50$-th epochs, respectively. The proposed model is trained with the Adam optimizer in a total of $60$ epochs. The size of the input image is $256 \times 128$. For data augmentation, we perform random cropping,
random flipping and auto-argumentation~\cite{cubuk2018autoaugment}. Besides, $m$ in in Eq.~\ref{eq01} and $\lambda$ in Eq.~\ref{eq06} are set as $0.2$ and $1.0$, respectively. 
 Particularly, we utilize the same setting for all experiments on all datasets in this paper.

\subsection{Comparison with State-of-the-art Methods}\label{sec:EXP-CUA}

We compare our proposed method with some state-of-the-art methods as reported in Table~\ref{tab01}, including QAConv$_{50}$~\cite{DBLP:conf/eccv/LiaoS20}, CBN~\cite{DBLP:conf/eccv/ZhuangWXZZWAT20}, SNR~\cite{DBLP:conf/cvpr/JinLZ0Z20}, OSNet~\cite{zhou2021learning} and M$^3$L~\cite{DBLP:conf/cvpr/ZhaoZYLLLS21}.
QAConv$_{50}$~\cite{DBLP:conf/eccv/LiaoS20} treats image matching as finding local correspondences in feature maps, and constructs query-adaptive convolution kernels on the fly to achieve local matching.
CBN~\cite{DBLP:conf/eccv/ZhuangWXZZWAT20} forces the images of all cameras to fall onto the same subspace, so that the distribution gap between any camera pair is largely shrunk.
SNR~\cite{DBLP:conf/cvpr/JinLZ0Z20} filters out style variations by instance normalization and  distill identity-relevant feature from the removed information and restitute it to the network to ensure high discrimination.
OSNet~\cite{zhou2021learning} is capable of learning omni-scale feature representation for person re-ID. When equipped with instance normalization via differentiable architecture search, OSNet becomes OSNet-AIN.
M$^3$L~\cite{DBLP:conf/cvpr/ZhaoZYLLLS21} utilizes a meta-learning strategy to simulate the train-test process of domain generalization for learning more generalizable models. 
As seen in Table~\ref{tab01}, our method outperforms all other methods on Rank-1 and mAP under different types of backbones. For example, in the ``D+C+Ms$\rightarrow$Ma'' task, our method increases M$^3$L by $+3.0\%$ ($55.5$ vs. $52.5$) and $+2.0\%$ ($80.3$ vs. $78.3$)  on mAP and Rank-1 when using IBN-Net50 as the backbone. This is mainly because our method mitigates the discrepancy between multiple source domains and the target domain. CBN takes into account the differences between different cameras, but does not deal with the different source domains. Both SNR and OSNet introduce normalization to improve the generalization of the model, but they are not effective in removing styles from multiple source domains, and feature distributions of the unseen target domains are also not well aligned. Our method not only improves the discrimination of the model, but also learn the domain-invariant feature representation and effectively deal with the domain-shift problem. In the end, we prove that our method has the potential to improve the generalization capability.

\begin{table}[htbp]
  \centering
  \caption{Comparison with the SOTA methods. ``D+C+Ms$\rightarrow$Ma'' denotes that the model is trained on DukeMTMC-reID (D), MSMT17 (Ms) and CUHK03 (C), and the model is tested on Market1501 (Ma). The \textbf{bold} is the best result.}
    \begin{tabular}{l|cc|cc}
    \toprule
    \multicolumn{1}{c|}{\multirow{2}[1]{*}{Method}}  & \multicolumn{2}{l|}{D+C+Ms$\rightarrow$Ma} & \multicolumn{2}{l}{Ma+C+Ms$\rightarrow$D} \\
\cmidrule{2-5}          & mAP   & Rank-1 & mAP   & Rank-1 \\
    \midrule
    QAConv(ResNet-50)~\cite{DBLP:conf/eccv/LiaoS20} & 39.5  & 68.6  & 43.4  & 64.9 \\
    CBN(ResNet-50)~\cite{DBLP:conf/eccv/ZhuangWXZZWAT20}   & 47.3  & 74.7  & 50.1  & 70.0 \\
    SNR(ResNet-50)~\cite{DBLP:conf/cvpr/JinLZ0Z20}   & 48.5  & 75.2  & 48.3  & 66.7 \\
    OSNet(OSNet)~\cite{zhou2021learning} & 44.2  & 72.5  & 47.0  & 65.2 \\
    OSNet(OSNet-IBN)~\cite{zhou2021learning} & 44.9  & 73.0  & 45.7  & 64.6 \\
    OSNet(OSNet-AIN)~\cite{zhou2021learning} & 45.8  & 73.3  & 47.2  & 65.6 \\
    M$^3$L(ResNet-50)~\cite{DBLP:conf/cvpr/ZhaoZYLLLS21} & 51.1  & 76.5  & 48.2  & 67.1 \\
    M$^3$L(IBN-Net50)~\cite{DBLP:conf/cvpr/ZhaoZYLLLS21} & 52.5  & 78.3  & 48.8  & 67.2 \\
    \midrule
    LDL(ResNet-50) ours & 51.3  & 77.6  & 52.6  & 71.9 \\
    LDL(IBN-Net50) ours & \textbf{55.5} & \textbf{80.3} & \textbf{55.1} & \textbf{72.3} \\
    \midrule
    \multicolumn{1}{c|}{\multirow{2}[1]{*}{Method}} & \multicolumn{2}{c|}{Ma+D+C$\rightarrow$Ms} & \multicolumn{2}{c}{Ma+D+Ms$\rightarrow$C} \\
\cmidrule{2-5}          & mAP   & Rank-1 & mAP   & Rank-1 \\
    \midrule
    QAConv(ResNet-50)~\cite{DBLP:conf/eccv/LiaoS20} & 10.0  & 29.9  & 19.2  & 22.9 \\
    CBN(ResNet-50)~\cite{DBLP:conf/eccv/ZhuangWXZZWAT20}   & 15.4  & 37.0  & 25.7  & 25.2 \\
    SNR(ResNet-50)~\cite{DBLP:conf/cvpr/JinLZ0Z20}   & 13.8  & 35.1  & 29.0  & 29.1 \\
    OSNet(OSNet)~\cite{zhou2021learning} & 12.6  & 33.2  & 23.3  & 23.9 \\
    OSNet(OSNet-IBN)~\cite{zhou2021learning} & 16.2  & 39.8  & 25.4  & 25.7 \\
    OSNet(OSNet-AIN)~\cite{zhou2021learning} & 16.2  & 40.2  & 27.1  & 27.4 \\
    M$^3$L(ResNet-50)~\cite{DBLP:conf/cvpr/ZhaoZYLLLS21} & 13.1  & 32.0  & 30.9  & 31.9 \\
    M$^3$L(IBN-Net50)~\cite{DBLP:conf/cvpr/ZhaoZYLLLS21} & 15.4  & 37.1  & 31.4  & 31.6 \\
    \midrule
    LDL(ResNet-50) ours & 18.4  & 43.9  & 30.9  & 31.6 \\
    LDL(IBN-Net50) ours & \textbf{21.6} & \textbf{49.4} &   \textbf{32.8}    &  \textbf{32.9}\\
    \bottomrule
    \end{tabular}%
  \label{tab01}%
\end{table}%

\subsection{Ablation Study}\label{sec:EXP-SS}
In this section, we conduct the ablation study to confirm the effectiveness of the proposed method, as shown in Table~\ref{tab02}. In this table, ``LDL-1'' indicates that the label distribution only contains the information of the relation between different classes, as described in Sec.~\ref{LDL-1}. ``LDL'' denotes the complete method, \ie, exploring the relation of different classes and reducing the domain gap across different domains as describe in Sec.~\ref{LDL-1} and Sec.~\ref{LDL-2}. As seen in this Table, when the label distribution is used for exploring the relation of different classes, the performance of the baseline can be obviously improved in all tasks. For example, in the ``D+C+Ms$\rightarrow$Ma'' task, using the ``LDL-1'' can improve the performance of the baseline by $+4.0\%$ ($48.9$ vs. $44.9$) on mAP. Therefore, this confirms the efficacy of  exploring the relation of different classes during training. Moreover, based on ``LDL-1'', combining the scheme of reducing the domain-shift across different domains can further enhance the result of the baseline, \eg, the performance of ``LDL'' is increased by $+2.1\%$ ($43.9$ vs. $41.8$) on Rank-1 when compared to ``LDL-1'' in the ``Ma+D+C$\rightarrow$Ms'' task, which owes to the scheme of reducing the domain gap. Hence, this ablation experiment validates that the proposed LDL method is beneficial for the domain generalization task.
\begin{table}[htbp]
  \centering
  \caption{Ablation study of the proposed LDL method using ResNet-50 as baseline. The \textbf{bold} is the best result.}
    \begin{tabular}{l|cccc}
    \toprule
    \multicolumn{1}{c|}{\multirow{2}[1]{*}{Moudle}} & mAP   & Rank-1 & Rank-5 & Rank-10 \\
\cmidrule{2-5}          & \multicolumn{4}{c}{D+C+Ms$\rightarrow$Ma} \\
    \midrule
    Baseline & 44.9  & 72.4  & 85.4  & 89.3 \\
    Baseline+LDL-1 & 48.9  & 75.9  & 87.6  & 91.3 \\
    Baseline+LDL & \textbf{51.3} & \textbf{77.6} & \textbf{88.5} & \textbf{92.0} \\
    \midrule
          & \multicolumn{4}{c}{Ma+C+Ms$\rightarrow$D} \\
    \midrule
    Baseline & 49.3  & 68.2  & 81.6  & 84.8 \\
    Baseline+LDL-1 & 51.1  & 70.3  & 82.3  & 85.2 \\
    Baseline+LDL & \textbf{52.6} & \textbf{71.9} & \textbf{82.8} & \textbf{86.2} \\
    \midrule
          & \multicolumn{4}{c}{Ma+D+C$\rightarrow$Ms} \\
    \midrule
    Baseline & 15.4  & 38.9  & 53.3  & 59.3 \\
    Baseline+LDL-1 & 17.0  & 41.8  & 55.9  & 61.6 \\
    Baseline+LDL & \textbf{18.4} & \textbf{43.9} & \textbf{58.0} & \textbf{63.8} \\
    \midrule
          & \multicolumn{4}{c}{Ma+D+Ms$\rightarrow$C} \\
    \midrule
    Baseline & 28.3  & 28.7  & 46.9  & 57.8 \\
    Baseline+LDL-1 & 28.6  & 29.4  & 47.5  & 56.6 \\
    Baseline+LDL & \textbf{30.9} & \textbf{31.6} & \textbf{50.6} & \textbf{60.3} \\
    \bottomrule
    \end{tabular}%
  \label{tab02}%
\end{table}%
 
 Besides, we also validate the effectiveness of our method on IBN-Net50 (\ie, using IBN-Net50 as the baseline), as reported in Table~\ref{tab03}. As seen in the table, our method is also effective in all tasks, \eg, mAP of the baseline is increased by $+4.8\%$ ($55.5$ vs. $50.7$) in the ``D+Ms+C$\rightarrow$Ma'' task, and the proposed LDL outperforms the baseline by $+5.2\%$ ($49.4$ vs. $44.2$) on Rank-1 in the ``Ma+D+C$\rightarrow$Ms'' task. As aforementioned, the proposed LDL method can not only explore the relation of different classes but also alleviate domain gap, thus it can bring the significant improvement of the performance.

\begin{table}[htbp]
  \centering
  \caption{Ablation study of the proposed LDL method using IBN-Net-50 as baseline.}
    \begin{tabular}{l|cccc}
    \toprule
    \multicolumn{1}{c|}{\multirow{2}[1]{*}{Method}} & \multicolumn{1}{l}{mAP} & \multicolumn{1}{l}{Rank-1} & \multicolumn{1}{l}{Rank-5} & \multicolumn{1}{l}{Rank-10} \\
\cmidrule{2-5}          & \multicolumn{4}{c}{D+Ms+C$\rightarrow$Ma} \\
    \midrule
    Baseline &  50.7     &   77.6    &   88.6    & 91.9  \\
    Baseline+LDL & 55.5  & 80.3  & 90.3  & 93.7 \\
    \midrule
          & \multicolumn{4}{c}{Ma+Ms+C$\rightarrow$D} \\
    \midrule
    Baseline &   51.8    &   69.8    &   81.6    & 85.1 \\
    Baseline+LDL & 55.1  & 72.3  & 84.1  & 87.2 \\
    \midrule
          & \multicolumn{4}{c}{Ma+D+C$\rightarrow$Ms} \\
    \midrule
    Baseline &  18.7     &  44.2     &   58.2    &  64.0 \\
    Baseline+LDL & 21.6  & 49.4  & 62.7  & 68.0 \\
    \midrule
          & \multicolumn{4}{c}{Ma+D+Ms$\rightarrow$C} \\
    \midrule
    Baseline &   25.0    &   22.6    &   43.9    & 54.7 \\
    Baseline+LDL &   32.8    &  32.9     &   52.0    & 62.5 \\
    \bottomrule
    \end{tabular}%
  \label{tab03}%
\end{table}%

Moreover, we also verify the efficacy of the proposed method when the unseen domains are small-scale datasets (\ie, PRID~\cite{DBLP:conf/scia/HirzerBRB11}, GRID~\cite{DBLP:journals/ijcv/LoyXG10}, VIPeR~\cite{DBLP:conf/eccv/GrayT08} and i-LIDs~\cite{DBLP:conf/bmvc/ZhengGX09}), and the model is trained on Marekt1501, DukeMTMC-reID, MSMT17 and CUHK03. Particularly, the performances of these small ReID datasets are evaluated on the average of 10 repeated random splits of gallery and probe sets. The experimental results are listed in Table~\ref{tab07}. It can be seen that our method can achieve a relatively large improvement on all small-scale datasets, \eg, mAP of the baseline is increased by $+8.2\%$ ($44.8$ vs. $36.6$) on the GRID. Our proposed method aligns the feature
space across different domains, so that it can achieve much better
performance. Such experimental results show the generalization potential of our model.

\begin{table}[htbp]
  \centering
  \caption{Experimental results on small-scale datasets when the model is trained on Market1501, DukeMTMC-reID, MSMT17 and CUHK03.}
    \begin{tabular}{p{5.665em}|cc|cc}
    \toprule
    \multirow{2}[1]{*}{Method} & mAP & \multicolumn{1}{p{4.085em}|}{Rank-1} & mAP & \multicolumn{1}{p{4.085em}}{Rank-1} \\
\cmidrule{2-5}    \multicolumn{1}{c|}{} & \multicolumn{2}{c|}{PRID} & \multicolumn{2}{c}{GRID} \\
    \midrule
    Baseline & 54.8  & 43.0  & 36.6  & 28.0  \\
    Baseline+LDL & 60.5  & 49.0  & 44.8  & 36.0  \\
    \midrule
    \multicolumn{1}{r|}{} & \multicolumn{2}{c|}{VIPeR} & \multicolumn{2}{c}{i-LIDs} \\
    \midrule
    Baseline & 63.2  & 52.5  & 80.3  & 71.7  \\
    Baseline+LDL   & 68.6  & 60.1  & 84.7  & 78.3  \\
    \bottomrule
    \end{tabular}%
  \label{tab07}%
\end{table}%

\subsection{Further Analysis}\label{sec:EXP-FA}
In this part, we conduct more experiments to further analyze the property of our method.

\begin{figure}
\centering
\subfigure[D+C+Ms$\rightarrow$Ma]{
\includegraphics[width=4.1cm]{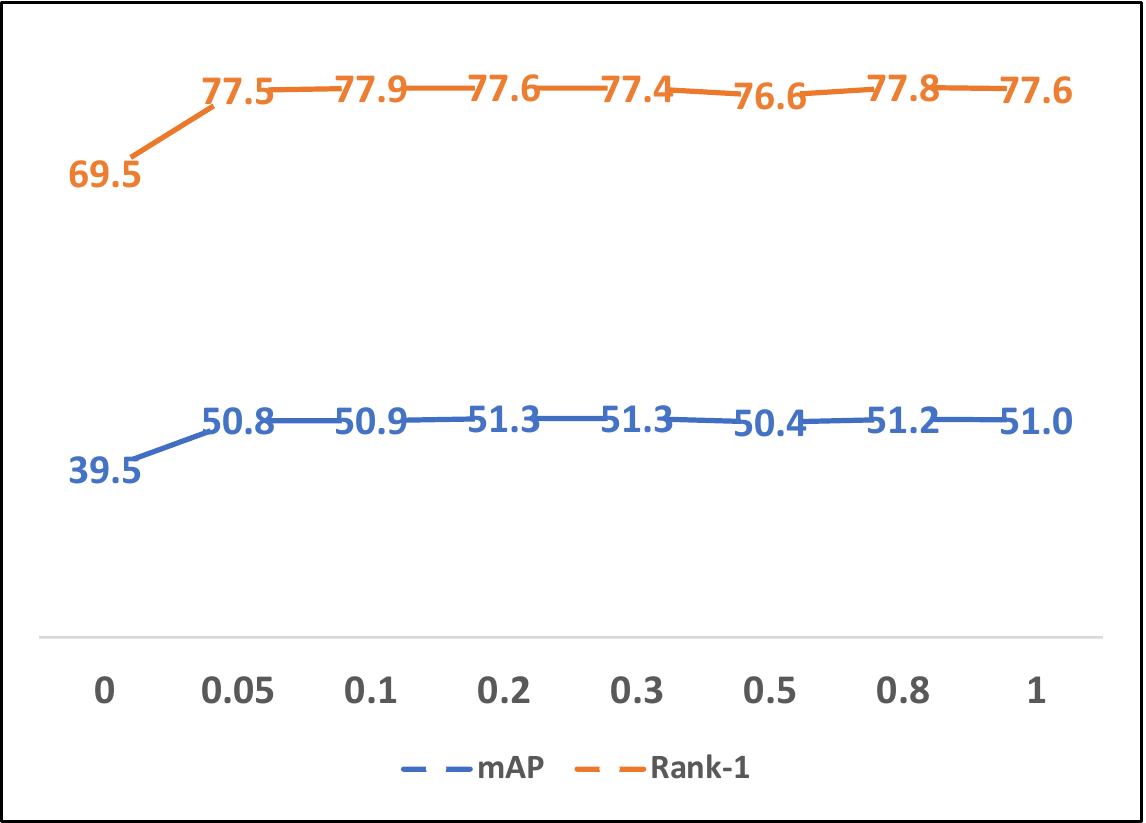}
}
\subfigure[Ma+Ms+C$\rightarrow$D]{
\includegraphics[width=4.13cm]{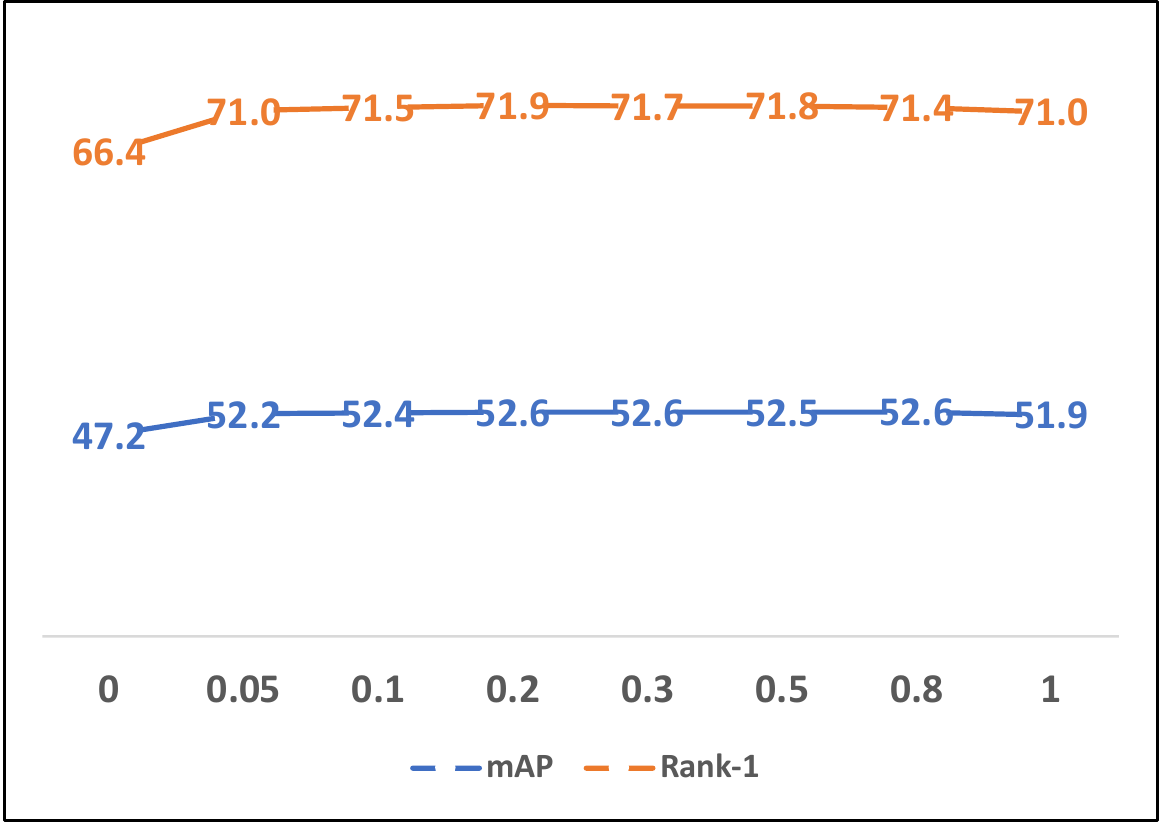}
}
\caption{Experimental results with different values of hyper-parameter $m$.}
\label{fig01}
\end{figure}

\begin{figure}
\centering
\subfigure[D+C+Ms$\rightarrow$Ma]{
\includegraphics[width=4.1cm]{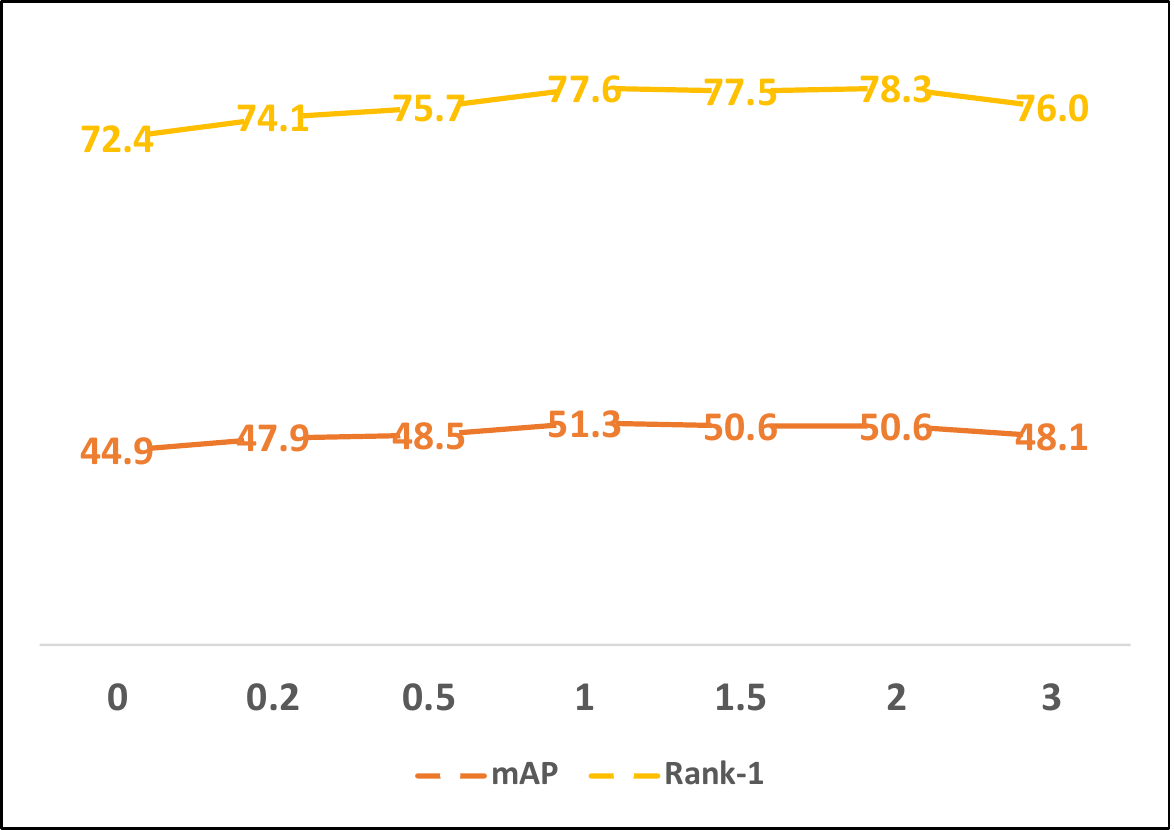}
}
\subfigure[Ma+Ms+C$\rightarrow$D]{
\includegraphics[width=4.04cm]{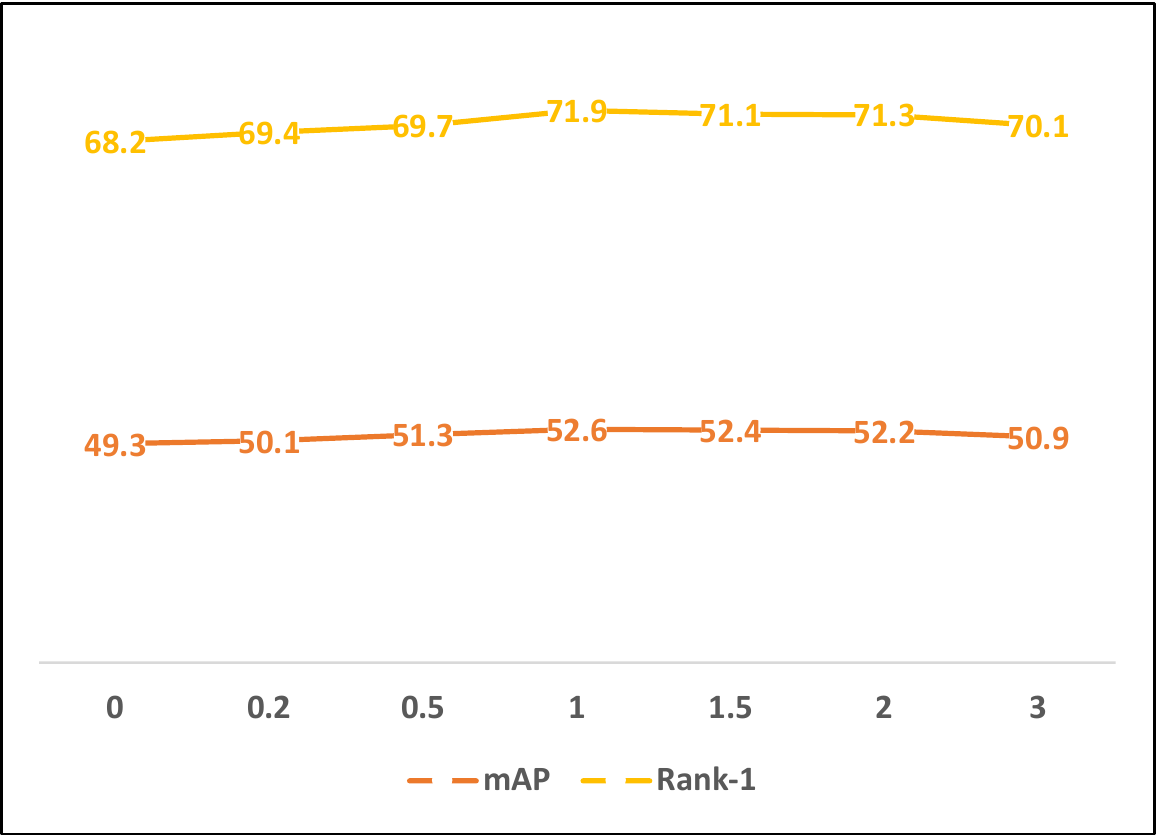}
}
\caption{Experimental results with different values of hyper-parameter $\lambda$.}
\label{fig02}
\end{figure}

\textbf{The sensitivity of the hyper-parameter.} In our method, there are two hyper-parameters \ie, $m$ and $\lambda$ in Eq.~\ref{eq01} and Eq.~\ref{eq06}. For the hyper-parameter $m$, we use it to update the matrix $\mathbf{M_T}$ after each iteration. In this experiment, we utilize the various $m$ to analyze the sensitivity of the hyper-parameter, as shown in Fig.~\ref{fig01}. If $m$ is set as $0$, it means the $\mathbf{M_T}$ is not updated during the whole training stage, thus the result is poor when compared to other settings. If $m$ is set as $1$, the performance slightly decreases in the ``Ma+Ms+C$\rightarrow$D'' task. According to the experiment, we set $m$ as $0.2$ in all experiments. Besides, for the hyper-parameter $\lambda$, it is used to trade off the proposed LDL and the conventional loss, and the experimental result is shown in Fig.~\ref{fig02}. As seen, when it is set as $1$, we can obtain relatively good result. If it is set as the larger or smaller value, the performance will slightly drop in all tasks. Hence, in our experiment, the hyper-parameter $\lambda$ is set as $1$.

\textbf{Further analysis of the proposed LDL.} To further confirm the effectiveness of the proposed LDL, we further conduct experiments in multiple different tasks, as reported in Table~\ref{tab04}. In this table, ``LDL-1'' denotes the proposed label distribution only explores the relation of different classes, as introduced in Sec.~\ref{LDL-1}. ``LDL-2'' represents that we further set the other classes in the same domain of the target class as $0$ to give more attention to the other domains, as described in Sec.~\ref{LDL-2}. ``LDL-3'' is the complete method as shown in Algorithm~\ref{al01}. Specifically, compared to ``LDL-2'', ``LDL-3'' further assigns the same attention to each other domain, except for the own domain, as described in Sec.~\ref{LDL-2}. As seen in Table~\ref{tab04}, when the label distribution is revised by the scheme of neglecting the classes of the own domain, the result can further be improved (\ie, ``LDL-2'' has better performance than ``LDL-1''). Moreover, ``LDL-3'' can further enhance the performance of ``LDL-2'', \eg, in the ``D+Ms+C$\rightarrow$Ma'' task, `LDL-3'' outperforms ``LDL-2'' by $+1.2\%$ ($51.3$ vs. $50.1$) on mAP. Therefore, the above results further validate the efficacy of the proposed method.
\begin{table}[htbp]
  \centering
  \caption{Evaluation of each component in the proposed LDL.}
    \begin{tabular}{l|cccc}
    \toprule
    \multicolumn{1}{c|}{\multirow{2}[1]{*}{Method}} & mAP   & Rank-1 & Rank-5 & Rank-10 \\
\cmidrule{2-5}          & \multicolumn{4}{c}{D+Ms+C$\rightarrow$Ma} \\
    \midrule
    LDL-1 & 48.9  & 75.9  & 87.6  & 91.3 \\
    LDL-2 & 50.1  & 76.7  & 88.1  & 92.0 \\
    LDL-3 & 51.3  & 77.6  & 88.5  & 92.0 \\
    \midrule
          & \multicolumn{4}{c}{Ma+Ms+C$\rightarrow$D} \\
    \midrule
    LDL-1 & 51.1  & 70.3  & 82.3  & 85.2 \\
    LDL-2 & 52.0  & 70.8  & 82.1  & 86.0 \\
    LDL-3 & 52.6  & 71.9  & 82.8  & 86.2 \\
    \midrule
          & \multicolumn{4}{c}{Ma+D+C$\rightarrow$Ms} \\
    \midrule
    LDL-1 & 17.0  & 41.8  & 55.9  & 61.6 \\
    LDL-2 & 17.9  & 42.8  & 57.1  & 63.3 \\
    LDL-3 & 18.4  & 43.9  & 58.0  & 63.8 \\
    \bottomrule
    \end{tabular}%
  \label{tab04}%
\end{table}%

\textbf{Evaluation on source domains.} We also evaluate our method on source domains as reported in Table~\ref{tab05}. As observed in this table, the proposed method (\ie, LDL) also outperforms the baseline model on source domains. For example, when the model is trained in the ``D+Ms+C$\rightarrow$Ma'' task, our method increases the result of MSMT17 (\ie, Ms) by $+2.5\%$ ($53.2$ vs. $50.7$) on mAP. This experiment shows that our method has a positive effect not only on the domain generalization task but also on the multi-domain person Re-ID task. The main reason is that the proposed label distribution learning can effectively exploit the information of each domain to boost the discrimination of the model by the class-relation-mining scheme and the domain-alignment scheme.

\begin{table}[htbp]
  \centering
  \caption{The experimental results on source domains.}
    \begin{tabular}{l|cc|cc|cc}
    \toprule
    \multicolumn{1}{c|}{Method} & mAP   & Rank-1 & mAP   & Rank-1 & mAP   & Rank-1 \\
    \midrule
          & \multicolumn{6}{c}{D+Ms+C$\rightarrow$Ma} \\
    \midrule
          & \multicolumn{2}{c|}{Test: D} & \multicolumn{2}{c|}{Test: Ms} & \multicolumn{2}{c}{Test: C} \\
    \midrule
    Baseline & 73.7  & 86.4  & 50.7  & 76.9  & 68.2  & 69.7 \\
    LDL   & 75.5  & 87.3  & 53.2  & 79.2  & 69.0  & 69.9 \\
    \midrule
          & \multicolumn{6}{c}{Ma+Ms+C$\rightarrow$D} \\
    \midrule
          & \multicolumn{2}{c|}{Test: Ma} & \multicolumn{2}{c|}{Test: Ms} & \multicolumn{2}{c}{Test: C} \\
    \midrule
    Baseline & 83.9  & 93.8  & 51.1  & 77.1  & 69.6  & 71.2 \\
    LDL   & 85.7  & 94.7  & 53.3  & 79.1  & 70.3  & 71.9 \\
    \midrule
          & \multicolumn{6}{c}{Ma+D+C$\rightarrow$Ms} \\
    \midrule
          & \multicolumn{2}{c|}{Test: Ma} & \multicolumn{2}{c|}{Test: D} & \multicolumn{2}{c}{Test: C} \\
    \midrule
    Baseline & 82.9  & 93.6  & 72.3  & 85.8  & 66.9  & 68.6 \\
    LDL   & 85.2  & 94.3  & 74.6  & 87.1  & 68.7  & 70.1 \\
    \midrule
          & \multicolumn{6}{c}{Ma+D+Ms$\rightarrow$C} \\
    \midrule
          & \multicolumn{2}{c|}{Test: Ma} & \multicolumn{2}{c|}{Test: D} & \multicolumn{2}{c}{Test: Ms} \\
    \midrule
    Baseline & 83.7  & 93.7  & 74.2  & 86.5  & 51.4  & 77.0 \\
    LDL   & 84.8  & 94.0  & 76.0  & 87.4  & 53.6  & 79.4 \\
    \bottomrule
    \end{tabular}%
  \label{tab05}%
\end{table}%

\begin{figure*}
\centering
\subfigure[D+C+Ms$\rightarrow$Ma (Baseline)]{
\includegraphics[width=4.1cm]{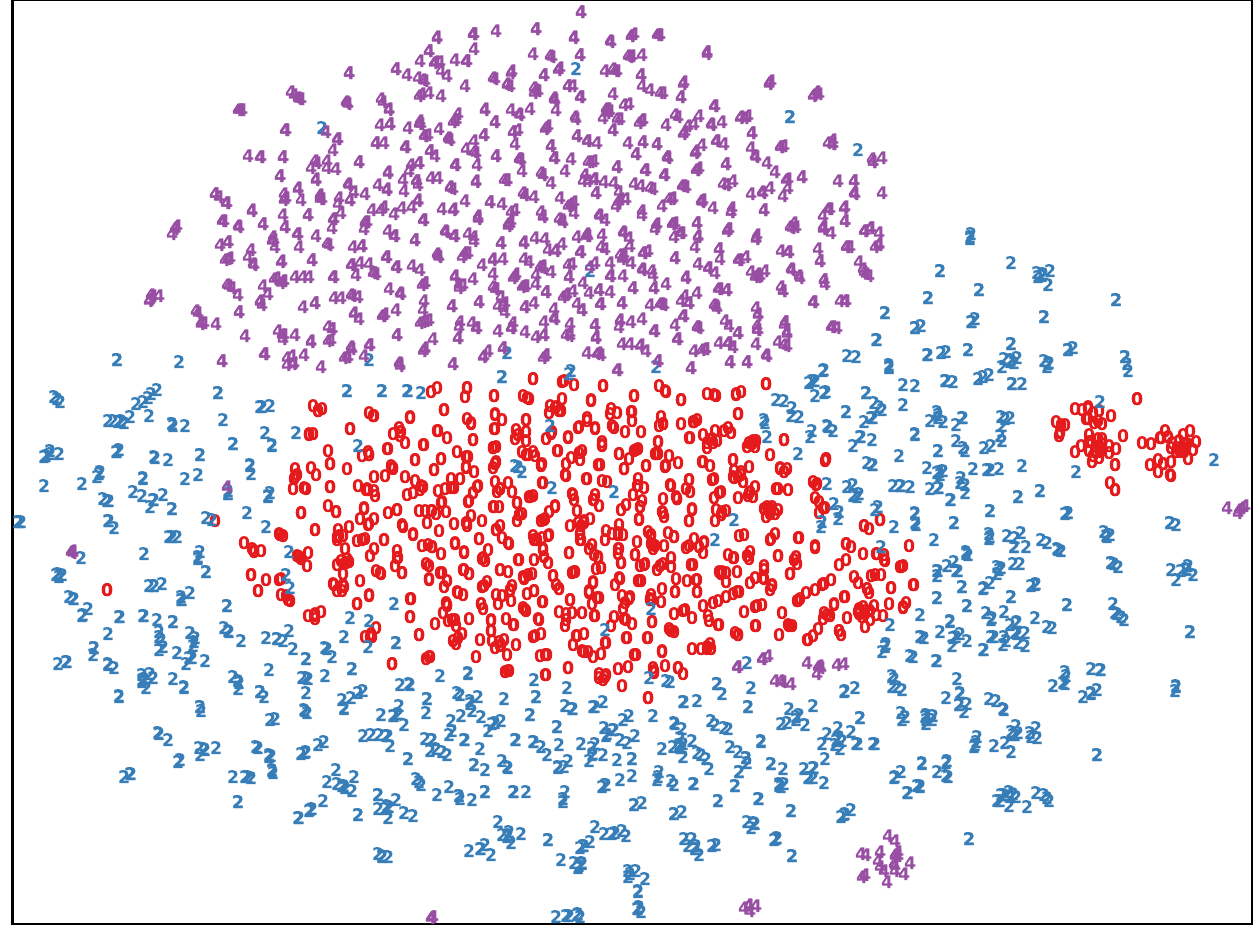}
}
\subfigure[Ma+Ms+C$\rightarrow$D (Baseline)]{
\includegraphics[width=4.04cm]{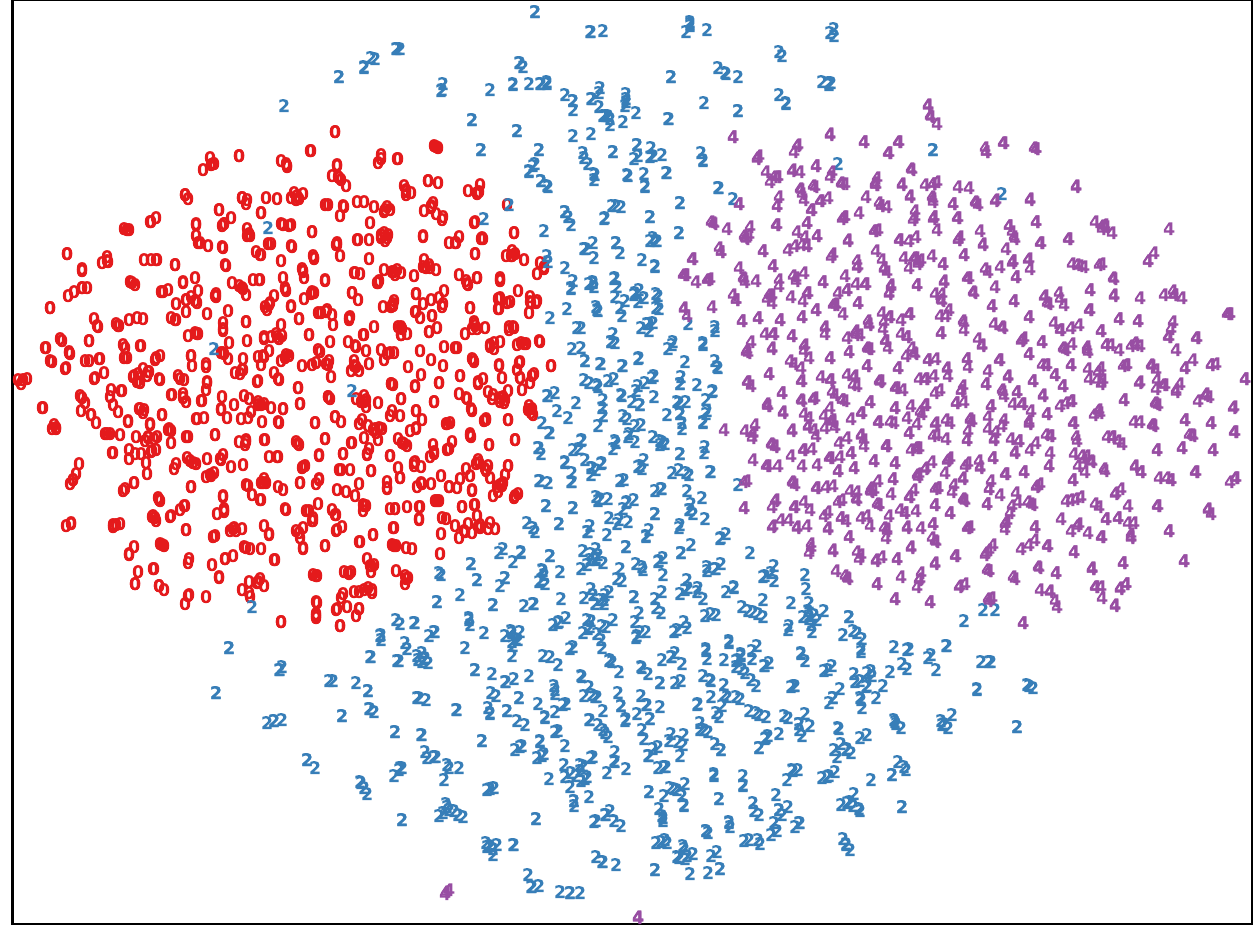}
}
\subfigure[Ma+D+C$\rightarrow$Ms (Baseline)]{
\includegraphics[width=4.04cm]{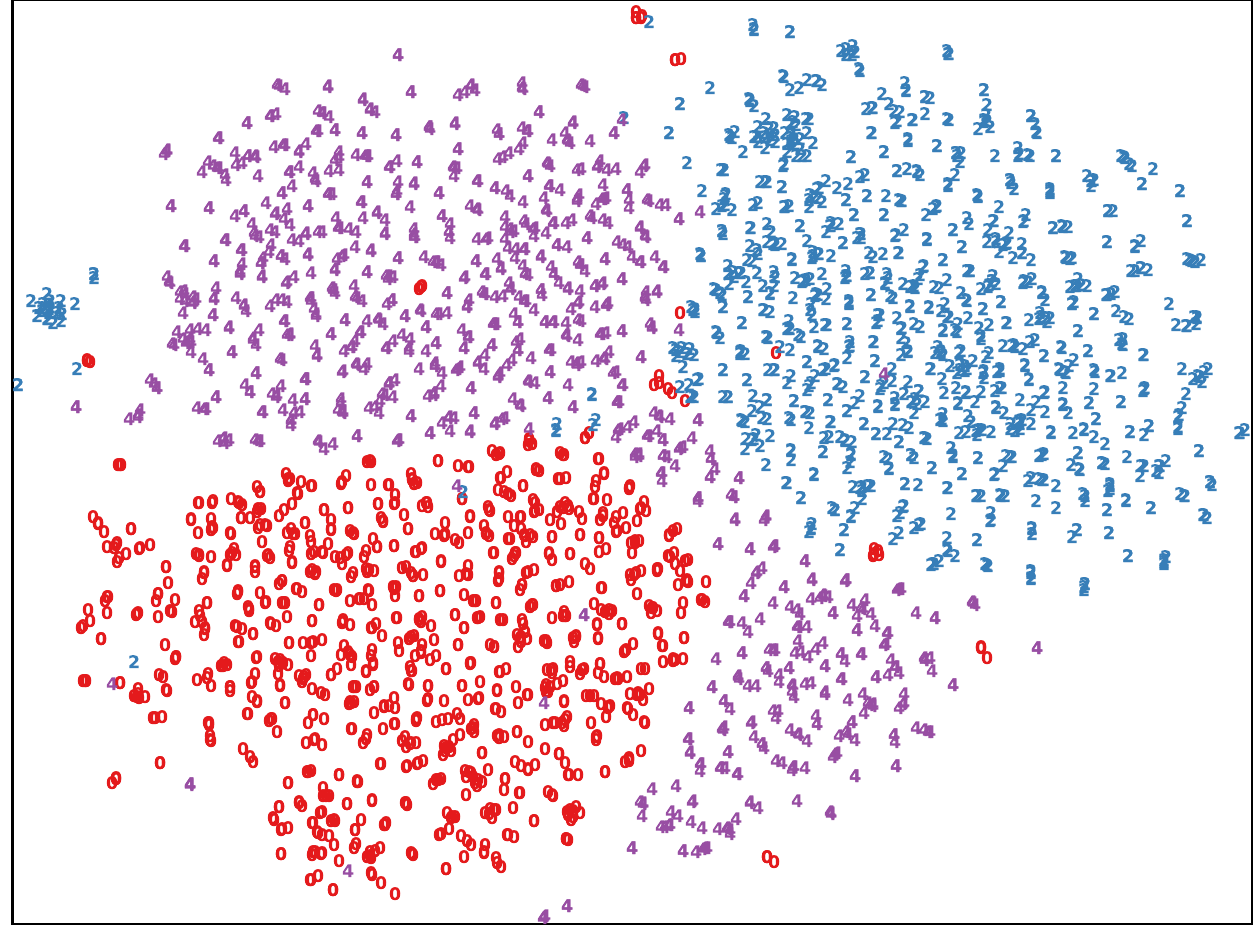}
}
\subfigure[Ma+D+Ms$\rightarrow$C (Baseline)]{
\includegraphics[width=4.04cm]{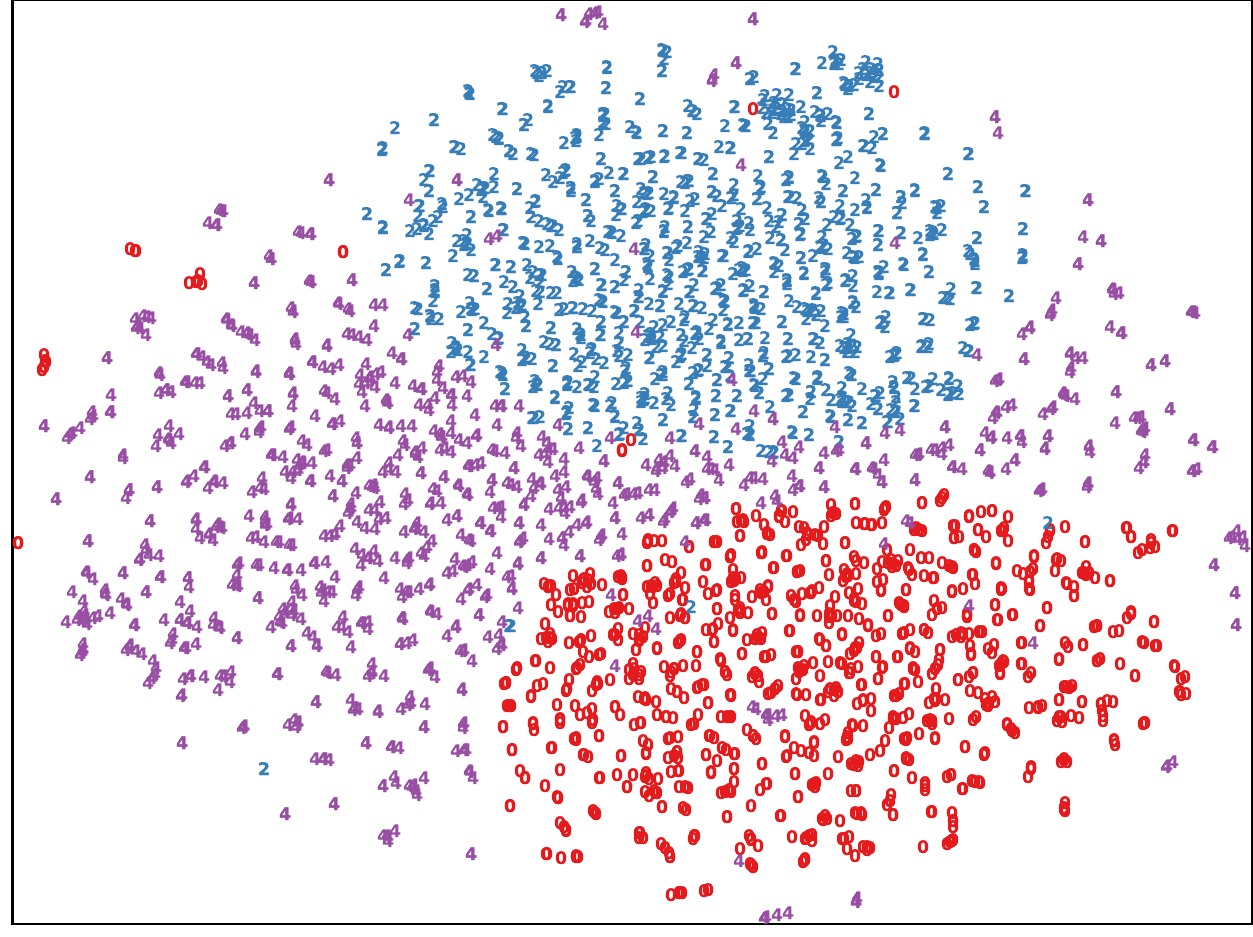}
}

\subfigure[D+C+Ms$\rightarrow$Ma (LDL)]{
\includegraphics[width=4.1cm]{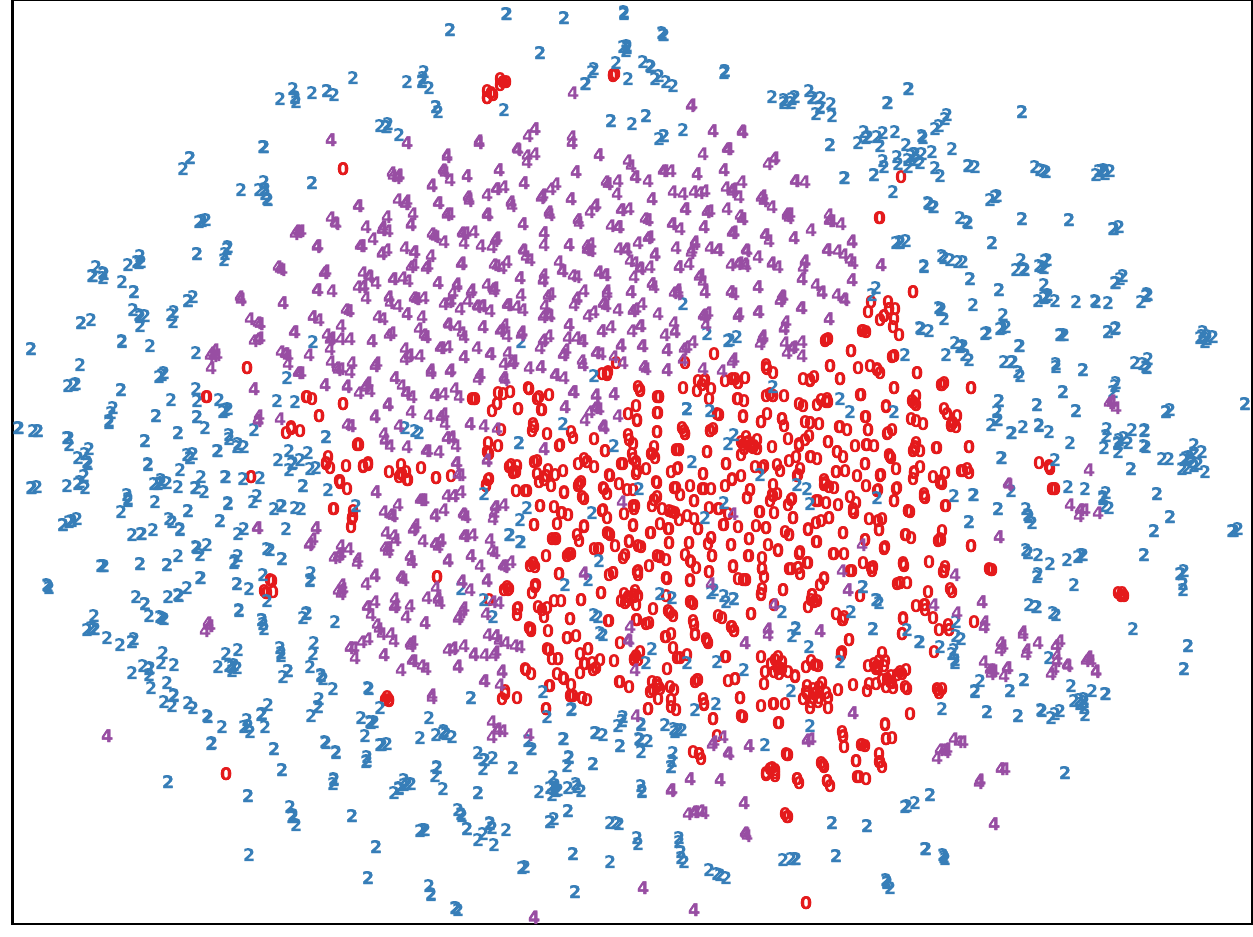}
}
\subfigure[Ma+Ms+C$\rightarrow$D (LDL)]{
\includegraphics[width=4.04cm]{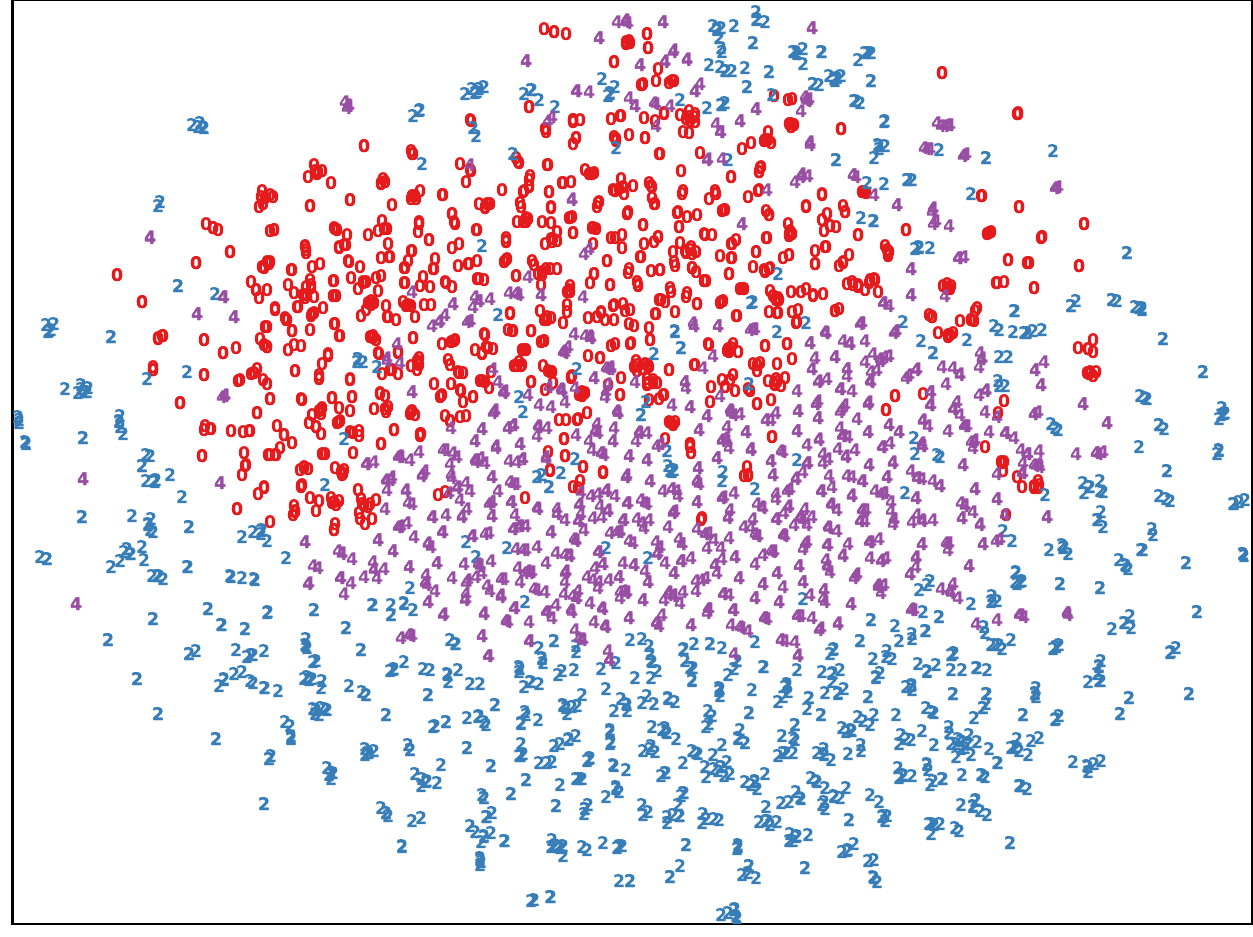}
}
\subfigure[Ma+D+C$\rightarrow$Ms (LDL)]{
\includegraphics[width=4.04cm]{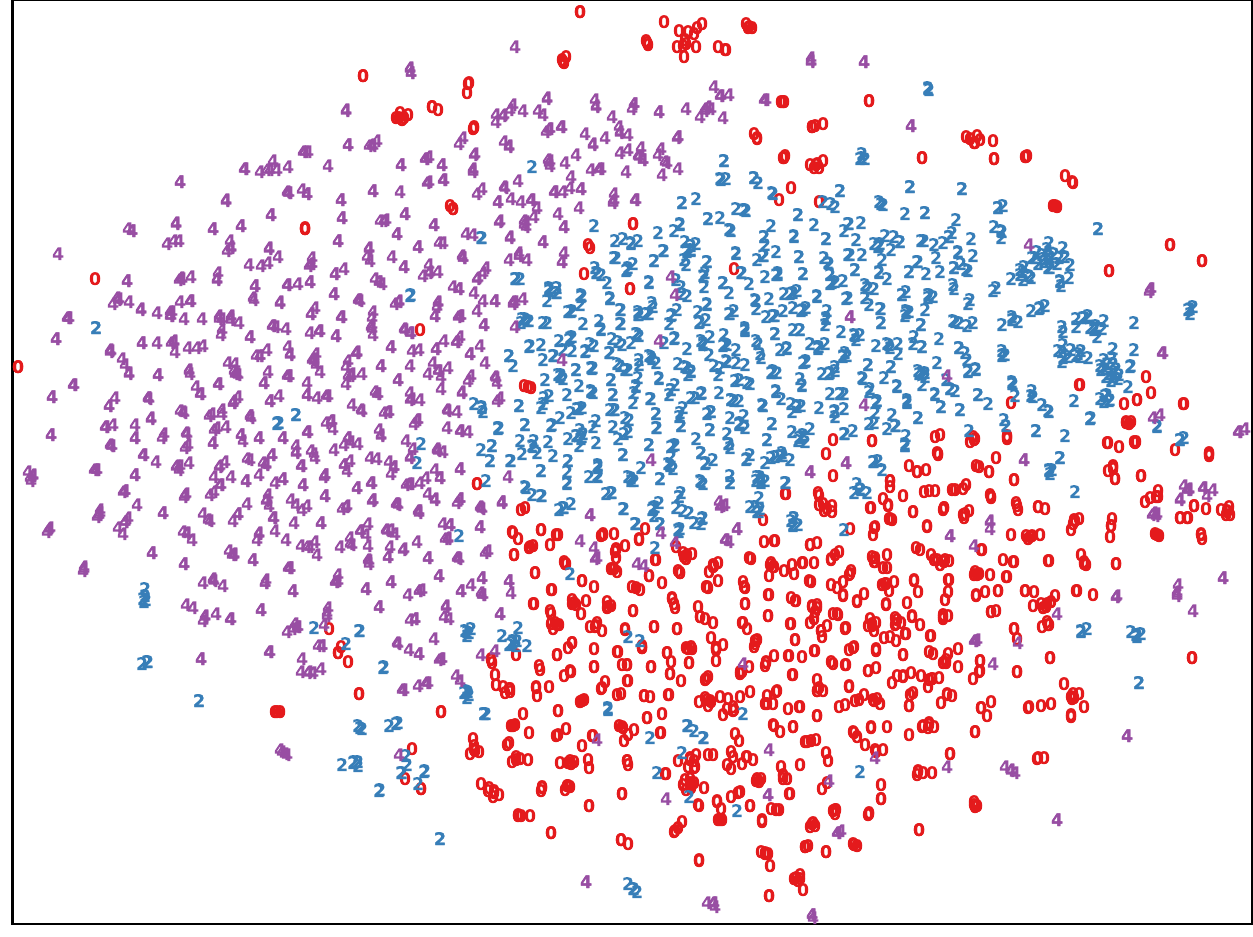}
}
\subfigure[Ma+D+Ms$\rightarrow$C (LDL)]{
\includegraphics[width=4.04cm]{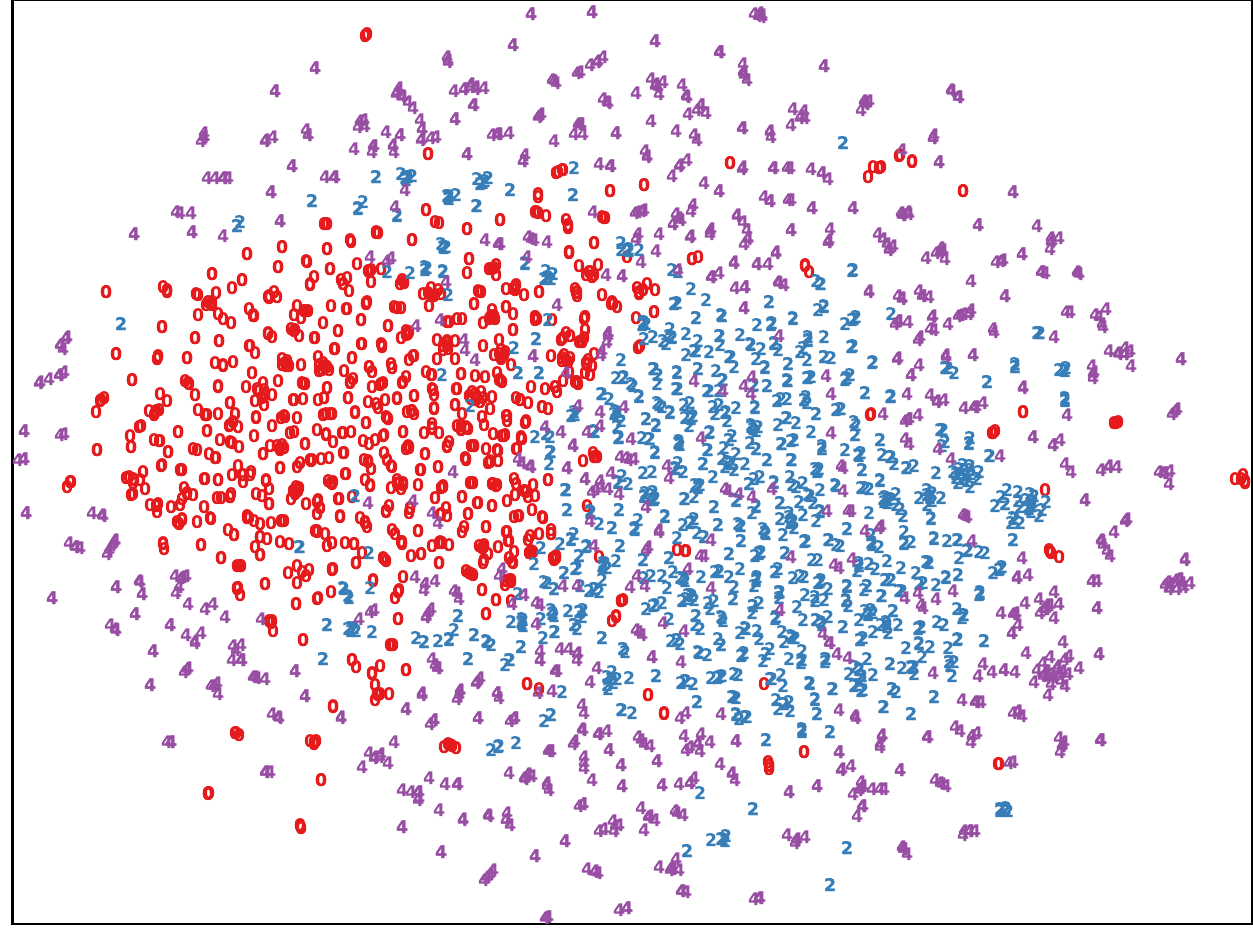}
}

\caption{Visualization of feature representation. The top line is the baseline, and the bottom is the proposed LDL method. Different colors denotes different domains. To conduct the fair comparison, we randomly select the same $1000$ samples from each domain to extract the feature in each task.}
\label{fig_vis}
\end{figure*}

\textbf{Visualization of feature representation.} In this part, we visualize the feature representation of our method and the baseline, as shown in Fig.~\ref{fig_vis}. As seen, each domain scatter in different regions in the visualization of the baseline, while our method tends to mix different domains into the same space. For example, in Fig.~\ref{fig_vis} (b), the red and the purple are almost non-overlapping, but in in Fig.~\ref{fig_vis} (f), the red and the purple are close. This experiment illustrates the proposed method can indeed mitigate the domain gap across different domains, which is consistent with the propose theoretical analysis in Sec.~\ref{T_A}.

\textbf{Experimental results of the label assignment.} To further reveal the property of our method, we display the similarity between a class and the other domains (\ie, all domains excluding the domain that the class belongs to). As described in Sec.~\ref{LDL-2}, the similarity is the averaged value of the corresponding domain in $\mathbf{M_T}$. The results are given in Table~\ref{tab06}, where ``OD-1'' and ``OD-2'' are the similarity between a class and other two domains, and the ``Diff'' denotes the difference between ``OD-1'' and ``OD-2''. For example, in the ``Market'' part, ``OD-1'' or ``OD-2'' represents the similarity between a class on Market1501 and DukeMTMC-reID or CUHK03. Particularly, the results are obtained after finishing the model training. In Table~\ref{tab06}, we randomly select three classes for each domain to show the results. There are two observations as follows: 1) Our proposed method enables classes to have higher similarity to other domains, which confirms that the proposed LDL gives more attention to other domains; 2) The ``Diff'' of the proposed method is smaller than the baseline, hence the proposed LDL focuses on the other domains more evenly when compared to the baseline. The above observations together deeply validates that our method can effectively alleviate the data-distribution discrepancy across different domains.
\begin{table}[htbp]
  \centering
  \caption{Experimental results ($\times 10^{-5}$) of the label assignment in the ``Ma+D+C$\rightarrow$Ms'' task.}
    \begin{tabular}{c|cc|c|cc|c}
    \toprule
    \multirow{2}[4]{*}{} & \multicolumn{3}{c|}{LDL} & \multicolumn{3}{c}{Baseline} \\
\cmidrule{2-7}          & OD-1     & OD-2     & Diff & OD-1     & OD-2     & Diff \\
    \midrule
    \multicolumn{1}{c|}{\multirow{3}[2]{*}{Ma}} & 5.13  & 5.28  & 0.15  & 4.22  & 4.02  & 0.20 \\
          & 5.50  & 5.45  & 0.05  & 4.42  & 3.97  & 0.45 \\
          & 5.60  & 5.63  & 0.03  & 3.69  & 3.22  & 0.47 \\
    \midrule
    \multicolumn{1}{c|}{\multirow{3}[2]{*}{D}} & 6.60  & 6.67  & 0.07  & 4.12  & 3.87  & 0.25 \\
          & 5.42  & 5.16  & 0.26  & 4.37  & 3.82  & 0.55 \\
          & 5.33  & 5.28  & 0.05  & 4.20  & 3.74  & 0.46 \\
    \midrule
    \multicolumn{1}{c|}{\multirow{3}[2]{*}{C}} & 5.34  & 5.37  & 0.03  & 4.06  & 4.21  & 0.15 \\
          & 4.73  & 4.86  & 0.13  & 3.84  & 4.11  & 0.27 \\
          & 6.76  & 7.08  & 0.32  & 4.50  & 5.73  & 1.23 \\
    \bottomrule
    \end{tabular}%
  \label{tab06}
\end{table}%

\textbf{Comparison between label distribution and one-hot label.} In this part, we conduct the experiments to demonstrate the superiority of using distribution over the one-hot label. In this experiment, we aim to conduct the direct comparison between one-hot label~\footnote{It is worth noting that the label smoothing scheme is not used.} (\ie, classification loss) and label distribution (\ie, label distribution learning loss). Particularly, since our label distribution depends on the classification loss, if there is no classification loss, the label distribution cannot be generated. Therefore, to ensure the identity information of the label distribution, we set the own identity as 0.88 in each label distribution. For example, if an image belongs to the $2$-nd class, the $2$-nd class is set as 0.88, and other classes are set using our label distribution method.  One-hot label and label distribution are implemented by ``$L_{overall}=L_{cls}+L_{tri}$'' and ``$L_{overall}=L_{ldl}+L_{tri}$'', respectively. Experimental results are reported in Table~\ref{tabR01}. As seen, using distribution outperforms the one-hot label in all tasks, which confirms the superiority of using distribution over the one-hot label.

 \begin{table}[htbp]
  \centering
  \caption{Comparison between label distribution and one-hot label.}
    \label{tabR01}%
    \begin{tabular}{l|cc}
    \toprule
    \multicolumn{1}{c|}{~~~~~~Lable Type~~~~~~} & ~~~~Rank-1~~~~ & ~~~~mAP~~~~ \\
    \midrule
    \multicolumn{3}{c}{D+Ms+C$\rightarrow$Ma} \\
    \midrule
    One-hot  & 71.59 & 45.10 \\
    Distribution & \textbf{75.59} & \textbf{49.84} \\
    \midrule
    \multicolumn{3}{c}{Ma+Ms+C$\rightarrow$D} \\
    \midrule
    One-hot  & 67.55 & 48.65 \\
    Distribution & \textbf{69.17} & \textbf{50.32} \\
    \midrule
     \multicolumn{3}{c}{Ma+D+C$\rightarrow$Ms} \\
    \midrule
    One-hot & 37.35 & 14.62 \\
    Distribution & \textbf{39.56} & \textbf{16.37} \\
    \midrule
    \multicolumn{3}{c}{Ma+D+Ms$\rightarrow$C} \\
    \midrule
    One-hot  & 28.86 & 27.51 \\
    Distribution & \textbf{32.29} & \textbf{31.20} \\
    \bottomrule
    \end{tabular}%
\end{table}%

\textbf{Evaluation on the classification distribution and feature representations for the domain gap.} Considering that using the cosine distance based on features is convenient for evaluating the similarity of different classes,  we conduct an experiment that utilizes the cosine distance based on feature representations. Specifically, we employ a memory bank to save the features of all classes with the momentum update manner. After each epoch, we use these features to compute the similarity of different classes, thus we can obtain a ${C \times C}$ matrix that is similar to $\mathbf{M_T} \in \mathbb{R}^{C \times C}$ in our method. We then leverage the same method as ours to produce $\mathbf{M_L}$, and use the same loss function as ours to train the model. The experimental results are reported in Table~\ref{tab_r2_1}. As observed, for estimating the domain gap, using the classification distribution outperforms using feature representations with the cosine distance. Besides, compared with the classification distribution scheme, using the feature representations with the cosine distance needs to compute the similarity after each epoch, thus it brings a larger computation cost during training.

\begin{table}[h]
  \centering
  \caption{Experimental results of different domain-gap evaluations, \ie, comparison between using the classification distribution and using feature representations with the cosine distance.}
  \label{tab_r2_1}%
    \begin{tabular}{l|cccc}
    \toprule
    \multicolumn{1}{c|}{\multirow{2}[1]{*}{Method}} & mAP   & Rank-1 & Rank-5 & Rank-10 \\
\cmidrule{2-5}          & \multicolumn{4}{c}{D+Ms+C$\rightarrow$Ma} \\
    \midrule
    Feature representations & 48.36  & 75.21  & 87.38  & 90.93  \\
    Classification distribution & 51.30  & 77.55  & 88.51  & 92.04  \\
    \midrule
          & \multicolumn{4}{c}{Ma+Ms+C$\rightarrow$D} \\
    \midrule
    Feature representations & 51.13  & 70.42  & 82.05  & 86.04  \\
    Classification distribution & 52.57  & 71.86  & 82.76  & 86.18  \\
    \midrule
          & \multicolumn{4}{c}{Ma+D+C$\rightarrow$Ms} \\
    \midrule
    Feature representations & 17.14  & 42.30  & 56.45  & 62.51  \\
    Classification distribution & 18.38  & 43.92  & 57.96  & 63.80  \\
    \bottomrule
    \end{tabular}%
\end{table}%

\section{Conclusion}\label{s-conclusion}
In this paper, we aim to address the generalizable multi-source person Re-ID task via label distribution learning. Different from the existing methods, we propose a novel label distribution learning method to explore the relation of different classes and reduce the domain-shift between different domains, which can enhance the discrimination of the feature and boost the generalization capability of the model. Furthermore, we give the theoretical analysis to validate the efficacy of the proposed method, which can map all features into the same space, thus it can enforce the model to learn the domain-invariant feature representation. Extensive experiments on multiple datasets validate the efficacy of the proposed method.


%
%

\ifCLASSOPTIONcaptionsoff
  \newpage
\fi

\bibliographystyle{IEEEtran}
\bibliography{sigproc}

 \vfill


\end{document}